\documentclass{article} % For LaTeX2e
\usepackage{iclr_files/iclr2027_conference,times}

\usepackage{amsmath,amsfonts,bm}

\def\eqref#1{equation~\ref{#1}}
\def\1{\bm{1}}

\DeclareMathAlphabet{\mathsfit}{\encodingdefault}{\sfdefault}{m}{sl}
\SetMathAlphabet{\mathsfit}{bold}{\encodingdefault}{\sfdefault}{bx}{n}

\usepackage{hyperref}
\usepackage{url}

\usepackage{graphicx}
\usepackage{subcaption}
\usepackage{amsmath}
\usepackage[dvipsnames]{xcolor}
\usepackage{wrapfig}

\usepackage{multirow}
\usepackage{algpseudocode}
\usepackage{algorithm}
\usepackage{enumitem}
\usepackage[capitalize]{cleveref}
\usepackage{booktabs}       % professional-quality tables

\title{Beyond Sample Copying: Structural Memorization in Diffusion Models}

\author{
  Tim Kaiser \\
  Heinrich-Heine-University Düsseldorf \\
  \And
  Markus Kollmann \\
  Heinrich-Heine-University Düsseldorf \\
}

\iclrfinalcopy % Uncomment for camera-ready version, but NOT for submission.
\begin{document}

\maketitle

\begin{abstract}
    Diffusion models generalize well in practice. 
Paradoxically, an optimal diffusion model fully memorizes the training data and therefore fails to generalize, raising the question of what induces generalization in a real diffusion model.
We show that diffusion models progressively overfit the denoising training objective, creating a generalization gap between validation and training performance at intermediate noise levels.
In a fully analytic 2D toy model with a controlled denoising error, we trace this gap to the interaction between model error and the density of the data distribution's support. The optimal denoising flow field localizes sharply around individual training points, whereas model error suppresses exact recall of training points, yielding a smooth, generalizing flow field.
Finally, we examine how training-time overfitting manifests along inference trajectories. We find that predictions made from intermediate trajectory states occupy a distinct feature-space regime from predictions made from noised training and validation images. As training progresses and model size increases, these predictions develop greater relative affinity to the training data, despite the absolute similarity to the validation data not decreasing. Together, these findings show that the denoising objective and inference trajectories express structural memorization differently.
 
All code will be released upon publication.

\end{abstract}

\newcommand{\recgap}{rec-gap}
\newcommand{\x}{\boldsymbol{x}}
\newcommand{\y}{\boldsymbol{y}}
\newcommand{\e}{\boldsymbol{\varepsilon}}
\newcommand{\n}{\boldsymbol{n}}
\newcommand{\q}{\boldsymbol{q}}
\renewcommand{\r}{\boldsymbol{r}}

\section{Introduction}
The goal of generative modeling is to sample from an unknown data distribution given only a finite set of examples. Diffusion models and closely related flow models have achieved strong performance on continuous data, generating novel samples across image \citep{sohldickstein2015deep,ho2020denoising,karras2022edm,karras2024edmv2,song2021scorebased,kingma2023variational}, video \citep{ho2022video,wang2023laviehighqualityvideogeneration,blattmann2023stablevideodiffusionscaling,photorealistic,melnik2024videodiffusionmodelssurvey}, and audio \citep{kong2021diffwave,huang2023makeanaudiotexttoaudiogenerationpromptenhanced,huang2025impactiterativemaskbasedparallel,liu2023audioldmtexttoaudiogenerationlatent} domains, while achieving competitive results with autoregressive models for text \citep{popov2021gradtts,xu2025energybaseddiffusionlanguagemodels,ochs2024privatesynthetictextgeneration,yi2024diffusion}.

Despite their impressive ability to generate high-quality, novel samples across different data modalities, diffusion models are known to occasionally replicate training samples \citep{carlini2023extractingtrainingdatadiffusion,somepalli2023understandingmitigatingcopyingdiffusion,gu2025memorizationdiffusionmodels,ren2025unveilingmitigatingmemorizationtexttoimage}. Supporting this notion, Bayes-optimal diffusion models are unable to sample anything other than the training data \citep{scarvelis2025closedformdiffusionmodels,adaloglou2025swg, gao2024flowmatchingmodelsmemorize}, %raising the question of \textit{how diffusion models generalize rather than how they memorize}. 
Practical generalization must therefore arise from the fact that learned flow fields only approximate this optimum. Existing analyses describe this approximation as a smoothing of the optimal field \citep{farghly2025diffusionmodelsmanifoldhypothesis, achilli2024losingdimensionsgeometricmemorization, scarvelis2025closedformdiffusionmodels} and identify regularization, inductive bias, and training dynamics as factors that influence the balance between generalization and memorization \citep{kamb2025analytictheorycreativityconvolutional,shah2025doesgenerationrequirememorization,bonnaire2025diffusionmodelsdontmemorize}. 

Analysing the noise schedule of diffusion models reveals that memorization risk is not uniform across noise levels \citep{dodson2026calmendswildmiddle}. Instead, the intermediate noise levels are the critical regime for determining both memorization and generalization. %Crucially, memorization formed there can propagate into the final sample level. 
Memorization risk is not uniform in the input space either. Instead, better diffusion models selectively fit the score in some regions of the input space, while underfitting it in others \citep{song2025selectiveunderfittingdiffusionmodels}. Specifically, as models improve, their score approximation improves on the regions of input space the model is highly likely to see during training, while it worsens outside of those regions. %As sampling trajectories typically lie beyond the training regions, this creates a structural disconnect between training and inference.

Memorization is not limited to verbatim copying, either. Garnier-Brun et al. recently described a form of biased generalization in which test loss continues to improve while generated samples become increasingly proximate to the training data \citep{garnierbrun2026biasedgeneralizationdiffusionmodels}. This result shows that held-out loss and final-sample proximity capture different aspects of generalization. It also raises a more specific question: how is training-specific memorization on the denoising objective expressed along inference trajectories?

%Together, these results suggest that memorization in training and inference depends both on the noise level and on the region of input space visited by the denoiser. Whether predictions from inference trajectories resemble predictions from unseen data, training-associated predictions, or a distinct feature-space regime remains open.

\begin{figure}[h]
     \centering
     \includegraphics[width=\textwidth]{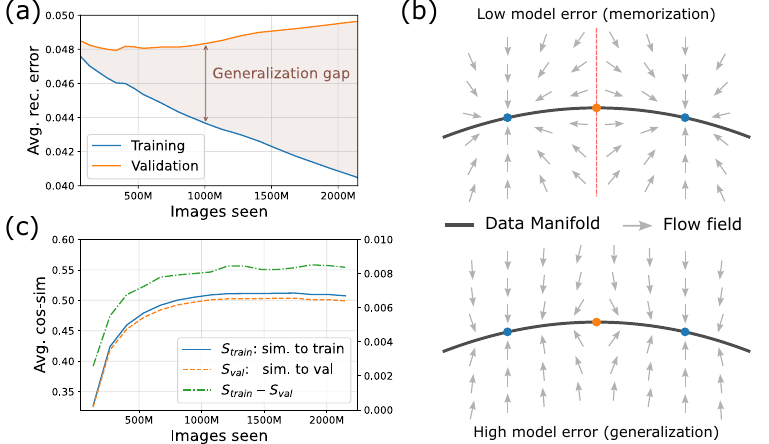}
     \caption{\textbf{Qualitative summary of contributions.} \textbf{(a)} State-of-the-art diffusion models exhibit a growing generalization gap: the difference in reconstruction error (denoising training objective) between training and validation data increases as training progresses. \textbf{(b)} This gap is modulated by model error: Increasing model error smoothes the learned flow field, suppressing the generalization gap. For low model error (top), the flow field localizes sharply around training points, while for high model error (bottom), it more smoothly approaches the data manifold. \textbf{(c)} Similarly, during inference, trajectories become increasingly biased towards the training data, relative to the validation data, despite increasing their absolute similarity to both.}
     \label{fig:qualitative_summary}
\end{figure}
\textbf{In this work}, we investigate this kind of structural memorization — a systematic difference in how diffusion models treat training versus unseen data — which persists independently of whether any given generated sample is a verbatim copy. %We show that overfitting at the denoiser level — visible when probing with noisy data samples — does not imply overfitting at the sampler level, and should therefore be treated as a related but separate quantity. %We show that memorization should be contrasted between training, a bias towards training data when reconstructing noisy data samples, and inference, an excessive similarity of generated samples to training samples. 
We trace this behaviour from the denoising objective, through a mechanistic toy model, to inference trajectories.

Our starting point is the denoising objective used by training, where we compare the reconstruction performance on noised training and validation examples at matched noise levels. This reveals a substantial generalization gap that increases with training time, model size, and decreasing dataset size. The gap is concentrated at intermediate noise levels and, at sufficiently long training, is accompanied by a decrease in validation performance while training performance continues to improve. 

By simulating an error-prone denoiser in a fully analytic 2D toy model, we explain how this gap can arise from the interplay between model error and the density of the data distribution's support. When model error is small, or the data manifold is sparsely supported, the learned flow field localizes sharply around individual training samples, leading to a large gap between training and validation performance. Larger model error or denser support smooths the field around the data support, thereby suppressing this gap.

Finally, turning to inference-level memorization, we define the relative feature-space affinity of predictions made from intermediate states of inference trajectories to the training reference set relative to the validation reference set. This measure differs from the reconstruction-based generalization gap, but it likewise probes differential treatment of training and validation data. We show that these predictions occupy a distinct feature-space regime from predictions made from directly noised training and validation images. %They show a relative training affinity that increases with the same broad factors -- training progression and model capacity -- that increase the earlier reconstruction gap. 
As training progresses and model capacity increases, these predictions develop greater relative affinity to the training data, while their absolute similarity to the validation data does not decrease. 

In summary, our analysis traces training-specific behaviour across levels of the diffusion process: from training objective denoising performance, through a mechanistic model of the learned flow field, to the feature-space geometry of inference trajectories.

\section{Better performing diffusion models show a larger generalization gap} \label{sec:l2-metric}
\begin{figure}[h]
    \centering
    \begin{subfigure}{0.34\textwidth}
        \centering
        \includegraphics[scale=0.4]{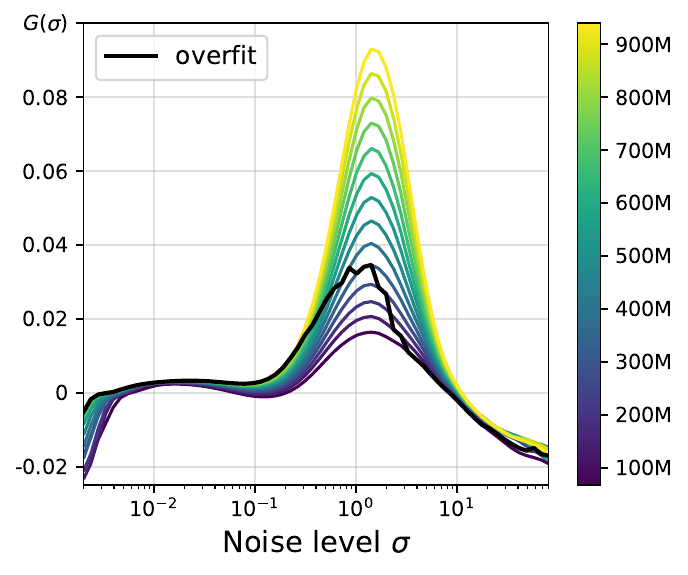}
        \caption{EDM2-XXL, IN-512}
    \end{subfigure} 
    % \begin{subfigure}{0.32\textwidth}
    %     \centering
    %     \includegraphics[scale=0.49]{imgs/residual-overfit-in512-edm2-xs.pdf}
    %     \caption{EDM2-XS, IN-512}
    % \end{subfigure}
    \begin{subfigure}{0.33\textwidth}
        \centering
        \includegraphics[scale=0.4]{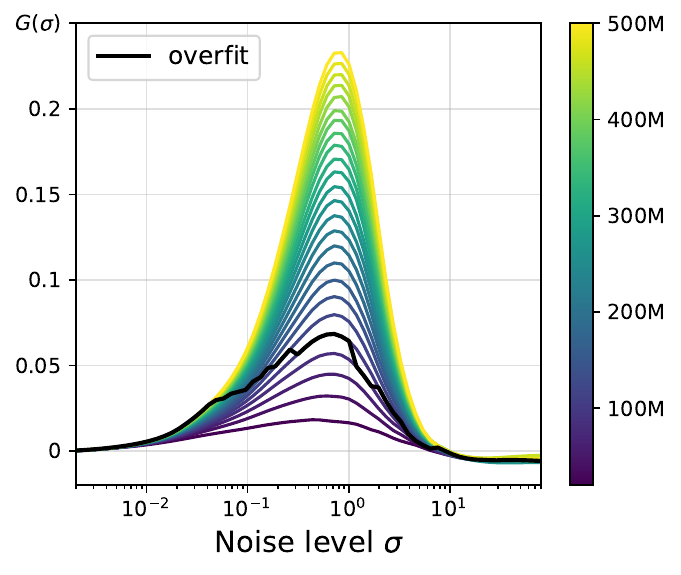}
        \caption{EDM, CIFAR-10}
    \end{subfigure}
    \begin{subfigure}{0.3\textwidth}
        \centering
        \includegraphics[scale=0.4]{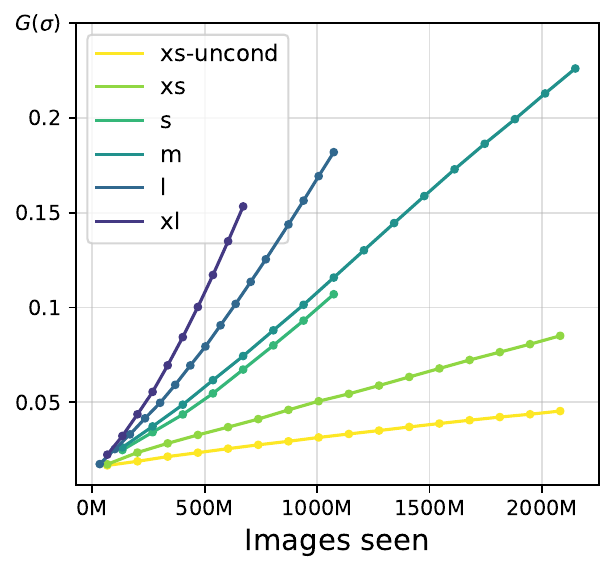}
        \caption{EDM2, IN-64}
    \end{subfigure}
    \caption{The relative generalization gap $G(\sigma)$ increases with images seen during training (colorbar in millions) and with model size in relation to dataset size. The black line indicates the beginning of overfitting (validation error starts increasing) for each noise level $\sigma$.}
    \label{fig:gen-gap}
\end{figure}
% To quantify overfitting, we compute the \textbf{relative generalization gap}
% \begin{equation} \label{eq:gen-gap}
%     G(\sigma) = \frac{M({\cal D}_\text{val}, \sigma) - M({\cal D}_\text{train}, \sigma)}{M({\cal D}_\text{train}, \sigma)},
% \end{equation}
% between a set of training points ${\cal D}_\text{train} = \{\y_i^\text{train}\}_{i=1}^N$ and a set of validation points ${\cal D}_\text{val} = \{\y_i^\text{val}\}_{i=1}^N$ at noise level $\sigma$.
% For now, we choose $M$ as the average L2-reconstruction error 
% \begin{equation} \label{eq:l2-error}
%     M_{L2}({\cal D}, \sigma) = \frac{1}{N}\sum_i \left\|\y_i^{\cal D} - \y(\x_i^{\cal D}, \sigma)\right\|_2^2,
% \end{equation}
% when reconstructing clean data points $\y_i$ from noisy data points $\x_i(\sigma) = \y_i + \sigma \n_i, \ \n_i \sim \mathcal{N}(0,I)$, using the target predictor $\y(\x,\sigma)$. 
To quantify the difference in denoising performance between training and validation data, we define the \textbf{relative generalization gap}
\begin{equation} \label{eq:gen-gap}
    G(\sigma) =
    \frac{L_{\mathrm{val}}(\sigma)-L_{\mathrm{train}}(\sigma)}
         {L_{\mathrm{train}}(\sigma)},
\end{equation}
where $L_{\mathrm{train}}$ and $L_{\mathrm{val}}$ denote the mean squared reconstruction errors at the noise level $\sigma$ on training and validation examples, respectively. For a dataset split ${\cal D}\in\{{\cal D}_\mathrm{train},{\cal D}_\mathrm{val}\}$, we define
\begin{equation} \label{eq:l2-error}
    L({\cal D},\sigma)
    = \frac{1}{|{\cal D}|}\sum_{i\in {\cal D}}
    \left\|\y_i-\y(\x_i,\sigma)\right\|_2^2,
\end{equation}
when reconstructing clean data points $\y_i$ from noisy data points $\x_i := \x_i(\sigma) = \y_i + \sigma \n_i, \ \n_i \sim \mathcal{N}(0,I)$, using the target predictor $\y(\cdot,\sigma)$. Denote $L_{\mathrm{train}}(\sigma) := L({\cal D}_{\mathrm{train}},\sigma)$ and $L_{\mathrm{val}}(\sigma) := L({\cal D}_{\mathrm{val}},\sigma)$.

%This directly probes a model's performance on the denoising task it is trained on, rather than comparing fully denoised samples, in contrast to Fréchet Distance-based generative metrics, such as FID. %We will further investigate this distinction in \cref{sec:inference}.
Perhaps unsurprisingly,
we find that well-performing denoiser models show a significant relative generalization gap between training and validation data (\cref{fig:gen-gap}), indicating that they exhibit different reconstruction performance on training and validation data for a given noise scale. This behavior is consistent across all models and datasets we tested. We use the pretrained EDM2 \citep{karras2024edmv2} models to carry out experiments on ImageNet and both pretrained and re-trained EDM \citep{karras2022edm} models for experiments on CIFAR-10 and CIFAR-100. The hyperparameters are as reported in \citep{karras2022edm,karras2024edmv2}, unless specified otherwise. Additional results for these models and datasets are available in the supplementary material.

%(\cref{fig:l2-supplement-in512}, \cref{fig:l2-supplement-in64})
We observe a positive correlation between the relative generalization gap and longer training times, larger model size, and smaller dataset size. This is consistent with findings of other memorization metrics, where the onset of memorization scales linearly with dataset size \citep{bonnaire2025diffusionmodelsdontmemorize,favero2025biggerisntmemorizingearly}. Additionally, we observe classical overfitting: during continued training, the validation reconstruction performance degrades (\cref{fig:gen-gap}, black line), while the training reconstruction performance improves. Crucially, both the relative generalization gap and the overfitting regime are most pronounced at intermediate noise levels and sharply drop outside of this range, consistent with previous findings about memorization being most pronounced at intermediate noise levels \citep{dodson2026calmendswildmiddle}. This raises the question of what causes the relative generalization gap to arise and how samples generated by standard diffusion sampling are affected by this.

\section{Model error causes the generalization gap} \label{sec:toy}
To understand the mechanism behind the noise-dependent relative generalization gap observed in \cref{sec:l2-metric}, we study a fully analytic 2D toy diffusion model, following \citep{adaloglou2025swg}. Consider a distribution of $N$ training points $\{\y^{train}_i\}_{i=1}^N$ and $N$ validation points $\{\y^\text{val}_i\}_{i=1}^N$ along a circle, alternating between each other. The shape of a circle was chosen for two reasons: it mimics the isotropy of high-dimensional data distributions, and the superposition of samples lies far outside the data manifold. %Training and validation points are sampled i.i.d. from this circle and split evenly.
The equidistant distribution of data points is chosen for simplicity. The results of this section are qualitatively unchanged for a random distribution on the circle. See supplementary material for more details. %\tim{Dataset density for random data acts strangely, why?.}
\begin{figure}[h]
     \centering
     \begin{subfigure}[b]{0.32\textwidth}
         \centering
         \includegraphics[scale=0.56]{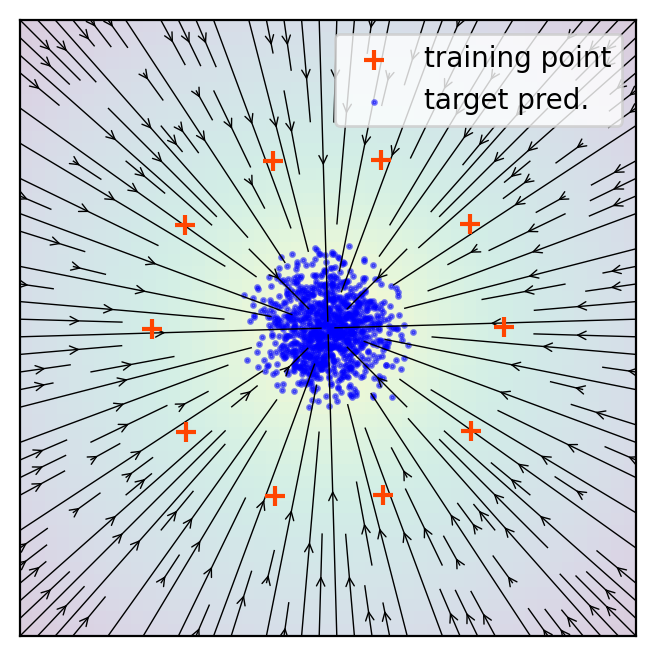}
         \caption{High noise}
     \end{subfigure} 
     \begin{subfigure}[b]{0.32\textwidth}
         \centering
         \includegraphics[scale=0.56]{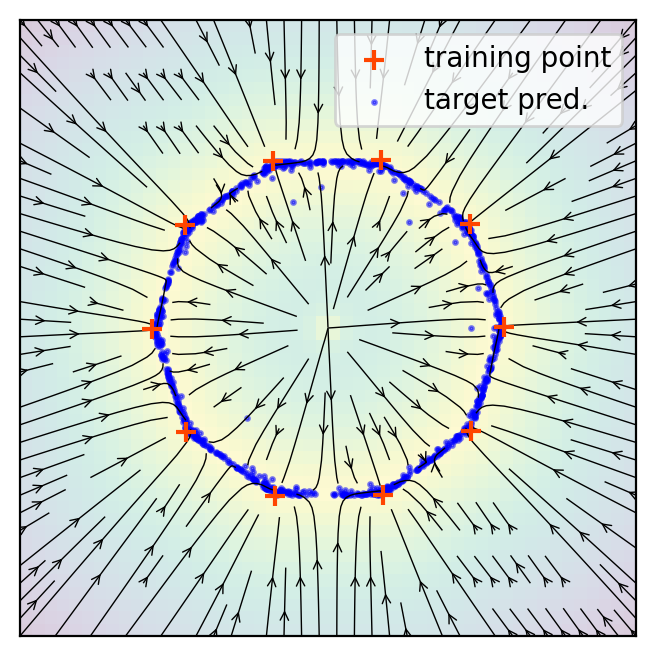}
         \caption{Intermediate noise}
     \end{subfigure}
     \begin{subfigure}[b]{0.32\textwidth}
         \centering
         \includegraphics[scale=0.56]{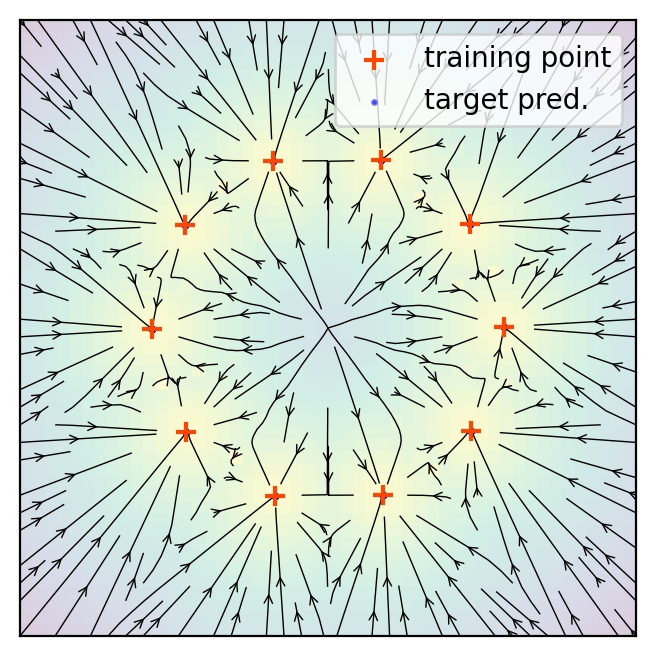}
         \caption{Low noise}
     \end{subfigure}
     \caption{Flow field lines of the optimal target predictor $\y$ at different noise levels $\sigma = 28, 2.8, 0.63$. Blue dots show predictions $\y^*(\x, \sigma)$ for $\x = \y + \sigma \n, \n \sim \mathcal{N}(0,I), \y \sim D_\text{train}$. Color indicates the magnitude of $\y^*(\x, \sigma) - \x$. Field geometry is \textbf{(a)} global, predictions tend towards the superposition of all training points, \textbf{(b)} moderately localized around training points, predictions approximate the data manifold, and \textbf{(c)} highly localized around each training point, predictions replicate the training points.}
     \label{fig:toy-geometry}
\end{figure}

Consider the time variable $t\in[0,1]$ and the current noise level $\sigma := \sigma(t)$, with $\sigma(0) = 0$ and $\sigma(1) \gg std(\bf y)$. We denote by $\x:= \x(\sigma(t))$ a point along a denoising trajectory. The \textbf{Bayes-optimal target predictor} is the posterior mean under the empirical training distribution \citep{karras2022edm}:
% \begin{equation} \label{eq:optimal-denoiser}
%     \y^*(\x, \sigma) 
%     := \int_{\mathbb{R}^d} \y P(\y|\x, \sigma^2)d\y 
%     = \int_{\mathbb{R}^d} \y \frac{\mathcal{N}(\x|\y, \sigma^2)P(\y)}{P(\x | \sigma^2)} d\y . 
% \end{equation}
% where ${\cal N}(.|.)$ denotes an isotropic normal distribution. For a finite set of training points, this takes the form 
\begin{align} \label{eq:optimal}
    \y^*(\x,\sigma)
    &= \frac{\sum_{i=1}^{N} \y_i^{\mathrm{train}}
        \,\mathcal{N}(\x\mid\y_i^{\mathrm{train}},\sigma^2 I)}
       {\sum_{i=1}^{N}
        \mathcal{N}(\x\mid\y_i^{\mathrm{train}},\sigma^2 I)}.
\end{align}
%Note that one can equivalently define the optimal noise predictor. 
%As a noisy sample $x$ approaches the data manifold, the $\y^*$ shifts from predicting the superposition of all training points at $\sigma(1)$ to being a projection onto the training points at $\sigma(0)$ (\cref{fig:toy-main}a).
This predictor defines a smooth vector field whose geometry depends on $\sigma$ (\cref{fig:toy-geometry}). In the high-noise limit, all training points contribute equally, giving
\begin{equation} \label{eq:superposition}
    \lim_{\sigma\to\infty}\y^*(\x,\sigma)
    = \frac{1}{N}\sum_{i=1}^{N}\y_i^{\mathrm{train}}.
\end{equation}
% This defines a smooth vector field over $\mathbb{R}^2$ whose geometry depends on $\sigma$ (\cref{fig:toy-geometry}), with limiting values
% \begin{equation} \label{eq:superposition}
%  \lim_{\sigma \rightarrow \infty} \y^*(\x, \sigma)  = \frac{1}{N}\sum_i \y^{train}_i,
% \end{equation}
% and
% \begin{equation} \label{eq:projection}
%     \lim_{\sigma \rightarrow 0} \y^*(\x, \sigma)  = \arg\min_{\y^{train}_i} \| \x - \y^{train}_i\|.
% \end{equation}
To see this, write 
\begin{equation}
    \y^*(\x,\sigma)
    = \frac{\sum_i \y_i^{\mathrm{train}}w_i(\x,\sigma)}
           {\sum_i w_i(\x,\sigma)}, %= \frac{\sum_i \y_i^{train}\cdot 1}{\sum_i 1} = \frac{1}{N}\sum_i \y_i^{train} 
\end{equation}
with
\begin{equation}
    w_i(\x,\sigma)
    := \exp\!\left(
        \frac{\x^{\top}\y_i^{\mathrm{train}}}{\sigma^2}
        - \frac{\|\y_i^{\mathrm{train}}\|_2^2}{2\sigma^2}
    \right).
\end{equation}
As $\sigma\to\infty$, $w_i(\x,\sigma)\to 1$ for every $i$, which yields \cref{eq:superposition}. In the low-noise limit, the posterior concentrates on the nearest training point, giving
\begin{equation} \label{eq:projection}
    \lim_{\sigma\to 0}\y^*(\x,\sigma)
    = \arg\min_{\y_i^{\mathrm{train}}}\|\x-\y_i^{\mathrm{train}}\|_2.
\end{equation}

For intermediate noise levels $\sigma$, a mixture of the above occurs, where noisy samples $\x$ are attracted by only their closest training samples. Sampling randomly initialized trajectories with this predictor yields perfect memorization of the training set. Note that the conditional optimal predictor is the same as $\y^*$, but the posterior mean includes only those training points satisfying the given condition.

\begin{figure}[h]
     \centering
     \begin{subfigure}[b]{0.32\textwidth}
         \centering
         \includegraphics[scale=0.4]{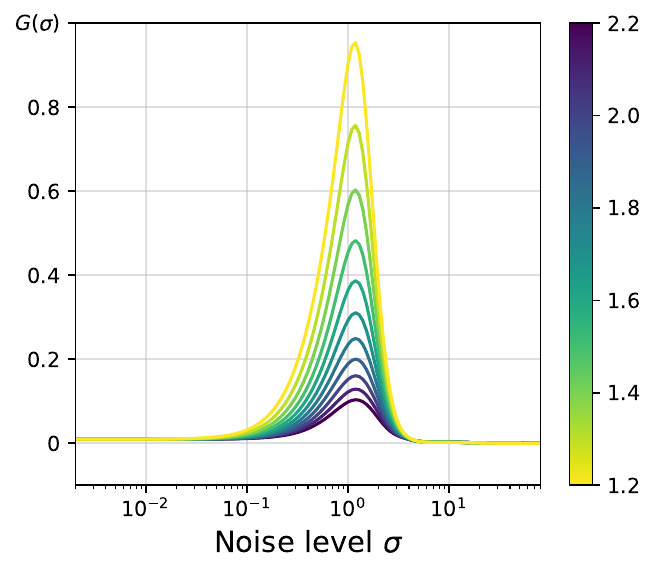}
         \caption{Error parameter $\delta$}
     \end{subfigure} 
     \begin{subfigure}[b]{0.32\textwidth}
         \centering
         \includegraphics[scale=0.4]{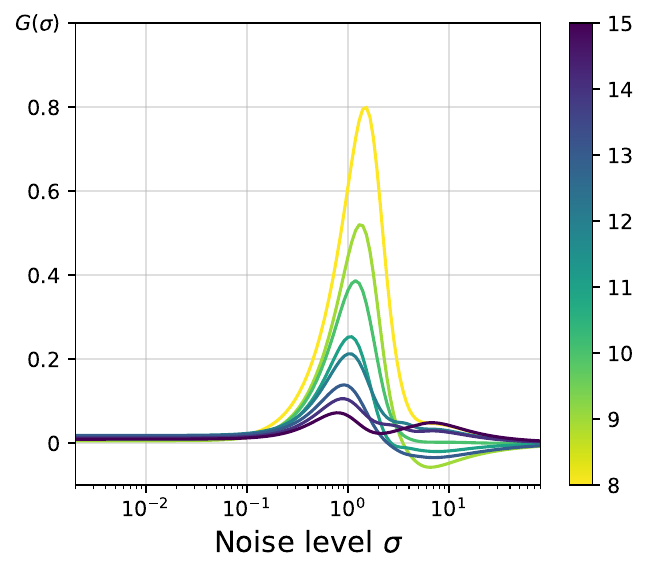}
         \caption{Training set size $N$}
     \end{subfigure}
     \begin{subfigure}[b]{0.32\textwidth}
         \centering
         \includegraphics[scale=0.4]{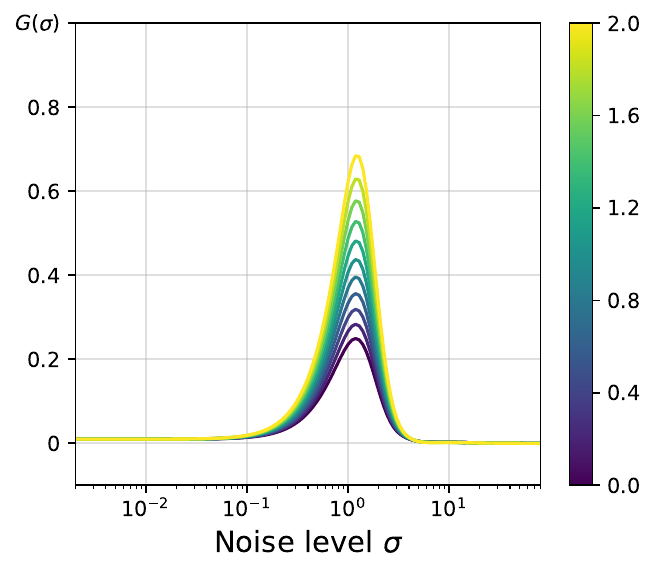}
         \caption{Guidance weight $w$}
     \end{subfigure}
     \caption{Relative generalization gap $G(\sigma)$ in our 2D toy model with circular data manifold \textbf{(a)} for different settings of the model error parameter $\delta$ (colorbar), \textbf{(b)} for different number of training points $N$ (colorbar), and \textbf{(c)} computed using a single point on the data manifold as the validation set, parametrized by its angle ($\sigma \approx 1.1$). The resulting gap is zero at training points, and largest between training points, with magnitude related inversely to $\delta$.}
     \label{fig:toy-gaps}
\end{figure}

To simulate real, erroneous diffusion models, we construct an \textbf{error-prone target predictor} $\y_\delta$ by substituting the real data distribution $P(\y)$ by an "uncertain" data distribution $P_\delta(\y) = \int_{\mathbb{R}^d} \mathcal{N}(\y|\y',\delta^2) P(\y') d\y'$. The true data points are randomly shifted by isotropic Gaussian noise with variance $\delta^2$, effectively broadening the hyperplane that the model fits. This construction yields the optimal denoiser for $\delta=0$, the "pass-through" property defined in \citep{lukoianov2025localityimagediffusionmodels}, and The smoothing effect of the model error parameter $\delta$ on the flow field connects to several characterizations of how practical diffusion models deviate from the optimal score \citep{li2024understandinggeneralizabilitydiffusionmodels,vastola2025generalizationvariancenoiseshapes,zhou2026smoothingscorefunctiongeneralization}, but provides a controlled setting in which the effect of this smoothing can be traced analytically. % \tim{and the hyperplane the samples land on for $\delta > 0$ approximates the real data hyperplane via interpolation between training samples, a behavior often associated with generalization in diffusion models [CITE]}. 
The corresponding posterior mean is given by
\begin{equation} \label{eq:error-prone}
    \y_\delta(x,\sigma)
=
\frac{\delta^2 \x + \sigma^2\, \y^*(\x,\tilde{\sigma})}
     {\tilde{\sigma}^2},
\qquad
\tilde{\sigma}^2 = \sigma^2 + \delta^2.
\end{equation}
Thus, model error both smooths the optimal predictor through the shift
$\sigma\mapsto\tilde\sigma$ and mixes it with the identity map through an affine combination \citep{adaloglou2025swg}. Whereas $\y^*$ can only reproduce training examples, $\y_\delta$ generalizes through random shifts induced by the initial noise of a trajectory. %We recover the optimal target predictor by $\y^* = \y_{\delta=0}$. 
It is interesting to see that despite its simplicity, the toy model can qualitatively reproduce the generalization gap behaviour of EDM/EDM2 trained on ImageNet data (\cref{fig:toy-gaps} vs \cref{fig:gen-gap}), using the same protocol as in \cref{sec:l2-metric}. Note that the toy model error parameter $\delta$ plays a role similar to training time in EDM/EDM2 in affecting the distance to the optimal denoiser.
%To visualize how $\delta$ affects the relative generalization gap along the data manifold, we consider a single point on the data manifold as the validation set and compute the generalization gap against it. The resulting gap is largest between training points, zero at training points, with magnitude related inversely to $\delta$ (\cref{fig:toy-gaps}c).
% \begin{figure}
%      \centering
%      \begin{subfigure}[b]{0.32\textwidth}
%          \centering
%          \includegraphics[scale=0.49]{imgs/d2.0e+00_s1.1e+00_contour.pdf}
%          \caption{$\delta=2.0$}
%      \end{subfigure} 
%      \begin{subfigure}[b]{0.32\textwidth}
%          \centering
%          \includegraphics[scale=0.49]{imgs/d1.6e+00_s1.1e+00_contour.pdf}
%          \caption{$\delta = 1.6$}
%      \end{subfigure}
%      \begin{subfigure}[b]{0.32\textwidth}
%          \centering
%          \includegraphics[scale=0.49]{imgs/d1.2e+00_s1.1e+00_contour.pdf}
%          \caption{$\delta = 1.2$}
%      \end{subfigure}
%      \caption{Filled contours showing the relative generalization gap $(E- E_{train})/E_{train}$ at intermediate noise ($\sigma \approx 1.1$) between the training points and each point in space, respectively. The black lines enclose the area where this gap is $0.5$ or less. As the model error $\delta$ decreases, this region shrinks until it no longer contains the validation points, indicating that the generalization gap to the validation set is larger than 0.5.}
%      \label{fig:toy-contours}
% \end{figure}

Analyzing the error-prone denoiser on noisy data points $\x = \y^{\cal D} + \sigma\n$, with $\cal D \in \{\text{train},\text{val}\}$, ${\n}\sim {\cal N}(0,I)$, explains why the gap emerges and why it scales with the model error $\delta$. We can write the \textit{prediction error} as
\begin{equation} \label{eq:decomposed}
\y_\delta(\x,\sigma) - \y = \frac{{{\sigma}^2}}{\tilde{\sigma}^2}\underbrace{(\y^*(\x,\tilde{\sigma}) - \y)}_{\text{optimal prediction error}} + \frac{\delta^2}{\tilde{\sigma}^2} {\sigma \n},
\end{equation}
which corresponds to the reconstruction error we used in \cref{eq:l2-error} to compute the relative generalization gap. 
%The prediction error (\cref{eq:decomposed}) only depends on $\y \sim D_\text{train/val}$ through the optimal prediction error. Therefore, a gap between the prediction error on training samples and validation samples can only arise when $\mathbb{E}_{\y \sim D_{\text{val}}}[||\y^* - \y||^2] > \mathbb{E}_{\y \sim D_{\text{train}}}[||\y^* - \y||^2]$. 
At intermediate noise levels ($\sigma_\text{min} \ll \sigma \ll \sigma_\text{max}$), the optimal flow field localizes around training points (\cref{fig:toy-geometry}b), causing a gap in the optimal prediction error in \cref{eq:decomposed} to arise between training and validation targets. This translates to a gap between the error-prone prediction errors (\cref{fig:toy-gaps}a) (the cross-term in the square of \cref{eq:decomposed} contributes about half as much). However, several factors can suppress this mechanism, leading to the characteristic shape of the gap observed in \cref{fig:gen-gap}:
At high noise levels ($\sigma \approx \sigma_\text{max} \gg \delta$), %the noisy samples $\x$ carry little information about the target $\y$, resulting in similar average errors between training and validation samples. Specifically, we have $\y_\delta = \y^* + {\cal O}(\delta^2/\sigma^2)$, and $\forall i: \mathcal{N}(\x|\y_i,\sigma^2) \approx N > 0$. Therefore, error-prone and optimal predictions collapse to $\y^*(\x, \sigma) \approx \frac{\sum_i \y^{train}_i N}{\sum_i N} = \frac{1}{N}\sum_i \y^{train}_i$. 
the error-prone prediction collapses to the superposition of all training samples:
\begin{equation}
    \y_\delta(\x, \sigma) = \y^*(\x, \sigma) + {\cal O}(\delta^2/\sigma^2) \overset{\cref{eq:superposition}}{\approx} \frac{1}{N}\sum_i \y_i^{train},
\end{equation}
resulting in similar errors to the training and validation samples.
At low noise levels or high model error ($\sigma^2/\delta^2 \ll 1$), we have $\y_\delta - \y = \sigma {\n} + {\cal O}(\sigma^2/\delta^2)$, which causes the error to drop with the noise level $\sigma$, regardless of whether $\y$ is sampled from the training or validation set.

The scale of the generalization gap is determined by the geometry around training points. Hence, it is also influenced by the density of training points on the manifold. The results in \cref{fig:gen-gap} suggest this, with the much smaller CIFAR datasets showing significantly higher levels of overfitting than the larger ImageNet dataset, especially when comparing models with a similar number of parameters. Similarly, theoretical analyses show that the memorization-generalization crossover depends on the ratio of data structure dimensionality to dataset size \citep{achilli2024losingdimensionsgeometricmemorization,buchanan2025edgememorizationdiffusionmodels}. Indeed, a denser distribution of training points on the manifold shows a significantly reduced gap in the toy model (\cref{fig:toy-gaps}b). Furthermore, this suggests that increasing the number of classes in class-conditional models (condition granularity) amplifies the gap, since in the toy model it's equivalent to reducing the dataset to specific subsets. Applying diffusion guidance \citep{karras2024guidingdiffusionmodelbad} to the toy model, it predicts a steady increase in the generalization gap as the guidance weight increases (\cref{fig:toy-gaps}c). We tested these predictions in \cref{sec:strong-condition,supp:guidance} against real diffusion models.

\section{Feature-Space Geometry of Diffusion Inference} \label{sec:inference}
In this section, we ask how the structural memorization we investigated in \cref{sec:l2-metric} and \cref{sec:toy} manifests in standard, class-conditional diffusion inference by examining the path inference trajectories take through space and the factors that influence it. Specifically, we want to answer the question of whether predictions made from inference trajectories are systematically biased toward the training set, resemble predictions made from unseen validation images, or occupy a distinct regime altogether. As intermediate trajectory states $\x(t)$ have no matched clean target, we cannot use the L2-reconstruction-based metric from \cref{sec:l2-metric} to analyse their relation to training and validation images. 

We therefore compare predictions generated from three query populations: noisy training images, noisy validation images, and intermediate states of standard class-conditional inference. Consider a query set that determines the model input, and a reference set that supplies clean images to which the resulting predictions are compared. Denote the query sets as $Q \in \{\text{train}, \text{val}, \text{traj}\}$. Each query image is passed through the model at noise level $\sigma$ to obtain a one-shot prediction $\y(\q,\sigma)$ of the corresponding clean image. For trajectory queries, $\sigma$ is the noise level of the intermediate state, while for noisy training and validation queries, $\sigma$ is chosen to match this same value. The reference can be either clean training or clean validation images, $R \in \{{\cal D}_\text{train}, {\cal D}_\text{val}\}$. 

To measure local similarity, we cannot use matched targets for noisy queries, and distributional metrics such as FID and FDD summarize differences between feature distributions and therefore do not directly measure whether individual query predictions are preferentially close to training rather than validation images. A viable alternative is to measure local similarity in feature space using cosine similarity. Specifically, consider the average cosine similarity between the predictions of each query image $\q \in Q$ and its $k$ nearest reference images $\r \in R$:
\begin{equation} \label{eq:similarity}
    S_k(Q, R; \sigma) = \frac{1}{|Q|}\sum_{\q \in Q} \frac{1}{k}\sum_{\r \in \text{kNN}(\y(\q, \sigma); R)} \cos(\phi(\y(\q, \sigma)), \phi(\r)), 
\end{equation}
where $\y(\cdot,\sigma)$ is a one-shot model prediction, $\phi$ is a feature map, and $\text{kNN}(\y(\q, \sigma); R)$ are the $k$ nearest neighbors of $\y(\q, \sigma)$ from $R$ in feature space and restricted to reference images from the same class as $\q$. We use the well-established DINOv2 feature extractor because of its reliability and robustness when evaluating generative models \cite{stein2023exposingflawsgenerativemodel}. When an image is part of both the query and the reference set, that image is excluded from the candidate neighbours, preventing the query’s own reference prediction from dominating the score.

\begin{figure}[h]
     \centering
     \begin{subfigure}[T]{0.32\textwidth}
         \centering
         \includegraphics[scale=0.38]{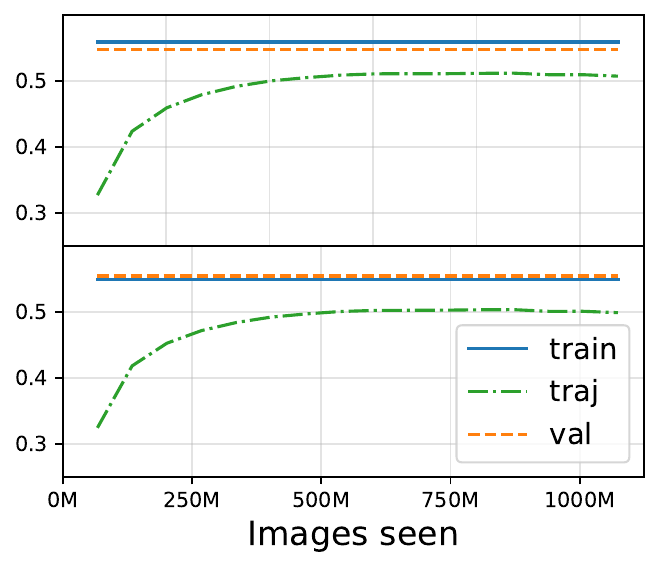}
         \caption{$S_{20}(Q, {\cal D}_\text{train/val}; \sigma)$}
     \end{subfigure} 
     \begin{subfigure}[T]{0.32\textwidth}
         \centering
         \includegraphics[scale=0.38]{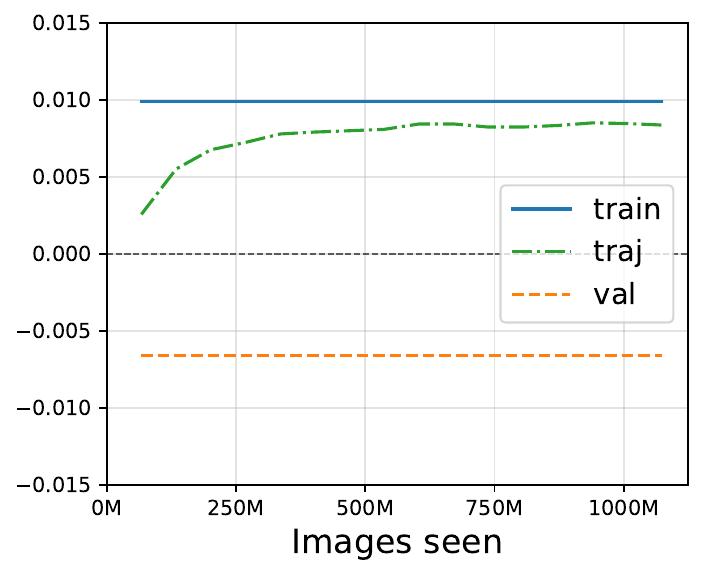}
         \caption{$B_{20}(Q; \sigma)$}
     \end{subfigure}
     \begin{subfigure}[T]{0.32\textwidth}
         \centering
         \includegraphics[scale=0.38]{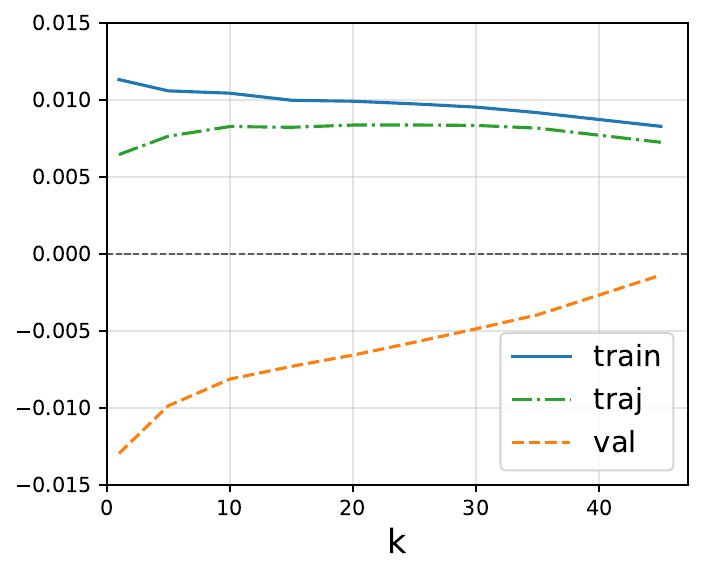}
         \caption{$B_k(Q; \sigma)$}
     \end{subfigure}
     \caption{\textbf{(a)} Average cosine similarity between predictions of query images $\q \in Q$ and clean images from the reference sets $\cal D_\text{train/val}$ (\cref{eq:similarity}). As training time increases, the absolute similarity of intermediate trajectory states (green line) to both reference sets increases. \textbf{(b)} So does their relative train-affinity (\cref{eq:affinity}, green line). \textbf{(c)} Their positive relative train-affinity is robust across choices of $k$. All plots show EDM2-S on ImageNet64.}
     \label{fig:q-v-ckpt}
\end{figure}

The hyperparameter $k$ defines how many nearest neighbors are used in \cref{eq:similarity}. For each $\y(\q,\sigma)$, small values of $k$ give greater weight to the closest reference points and are therefore more sensitive to memorized images or near-duplicates. Larger $k$ average over a wider local neighbourhood, making the statistic more representative of regional feature-space similarity. We use $k=20$, balancing sensitivity to individual near-neighbours against the variance of a single-neighbour statistic. The qualitative trends remain unchanged across the range of values we tested for $k$ (\cref{fig:q-v-ckpt}c, \cref{fig:in64-Q-v-sigma-k}).

Using these absolute similarity measures, we define the relative train-affinity of $Q$:
\begin{equation} \label{eq:affinity}
    B_k(Q; \sigma) = S_k(Q, {\cal D}_\text{train}; \sigma) - S_k(Q, {\cal D}_\text{val}; \sigma).
\end{equation}
If $B_k(Q; \sigma) > 0$, then model predictions from $Q$ are, on average, more similar to the training reference set than to the unseen validation reference set in feature space. For a full description of our experimental pipeline, see \cref{supp:reproducibility}.
\begin{figure}[h]
     \centering
     \begin{subfigure}[b]{0.32\textwidth}
         \centering
         \includegraphics[scale=0.38]{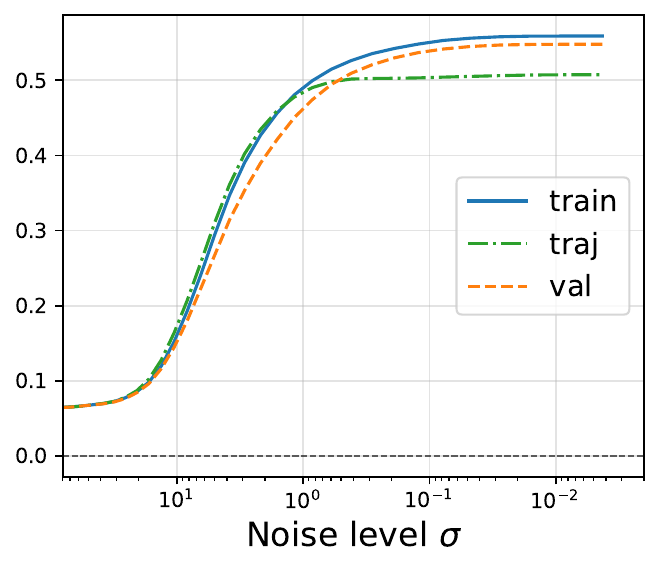}
         \caption{$S_{20}(Q, {\cal D}_\text{train};\sigma)$}
     \end{subfigure} 
     \begin{subfigure}[b]{0.32\textwidth}
         \centering
         \includegraphics[scale=0.38]{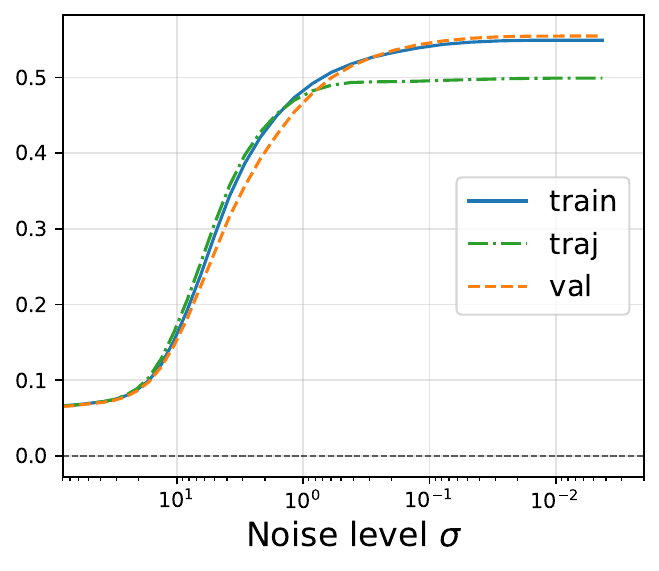}
         \caption{$S_{20}(Q, {\cal D}_\text{val};\sigma)$}
     \end{subfigure}
     \begin{subfigure}[b]{0.32\textwidth}
         \centering
         \includegraphics[scale=0.38]{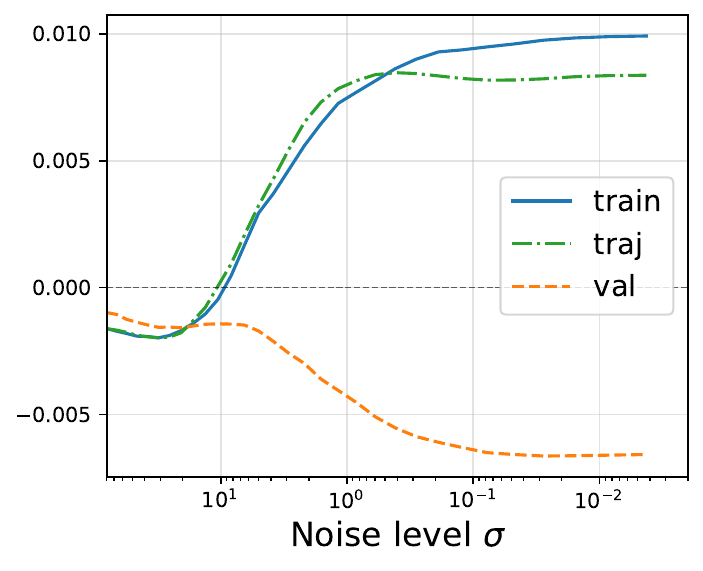}
         \caption{$B_{20}(Q; \sigma)$}
     \end{subfigure}
     \caption{\textbf{(a) and (b)} Average cosine similarity between predictions of query images $\q \in Q$ and clean images from the reference sets $\cal D_\text{train/val}$ (\cref{eq:similarity}). As the noise level $\sigma$ decreases, overall similarity to the reference sets increases. At low noise, both noisy training and noisy validation images as queries (blue and orange lines) exceed intermediate trajectory states (green line) in absolute similarity to both reference sets. \textbf{(c)} The relative train affinity (\cref{eq:affinity}) of intermediate trajectory states (green line) increases alongside the absolute similarities to the reference sets. All plots show EDM2-S on ImageNet64.}
     \label{fig:q-v-sigma}
\end{figure}
Similarity to the training reference set, $S_k(Q, {\cal D}_\text{train}; \sigma)$, measures how closely the predictions are associated with clean training images in feature space, whereas similarity to the validation reference set, $S_k(Q, {\cal D}_\text{val}; \sigma)$, provides the corresponding comparison for unseen images from the true data distribution.

\cref{fig:q-v-sigma}a and b show that, as soon as the noise level allows for discernible image features, intermediate trajectory states have lower absolute similarity to both reference sets than predictions from noised training or validation images. Assuming the reference sets sufficiently cover the data manifold and that DINOv2 similarity reflects proximity to it, this gap indicates that the trajectories cannot fully close the distance to the manifold of the true data distribution. Furthermore, trajectory-derived predictions exhibit a strong relative train-affinity (\cref{fig:q-v-sigma}c). This affinity is most pronounced at low noise, even though trajectory-derived predictions remain less similar to both reference sets in absolute terms. This provides feature-space evidence from inference trajectories that is consistent with the structural memorization identified by the paired L2 analysis, while measuring a different quantity.

This geometric separation extends the supervision-region perspective of Dodson et al. \cite{dodson2026calmendswildmiddle}: Our comparison shows that intermediate trajectory states are separated not only from the regions represented by noisy training queries, but also from the regions represented by directly noised validation queries. Thus, the relevant distinction here is not only between supervised and unsupervised inputs: trajectory states occupy a feature-space regime distinct from directly noised data from the entire data distribution.

With longer training, and likewise with larger model size, the absolute-similarity gap between trajectory-derived predictions and predictions from directly noised images progressively closes, closing the aforementioned distance to the manifold of the data distribution (\cref{fig:q-v-ckpt}a, \cref{fig:in64-Q-v-sigma}, \cref{fig:q-v-ckpt}). This improvement comes at the cost of a larger relative train-affinity (\cref{fig:q-v-ckpt}b), consistent with the increasing relative generalization gap between training and validation performance in \cref{sec:l2-metric} and \cref{sec:toy}. Because similarity to the validation reference set, $S_k(\text{traj}, {\cal D}_\text{val}; \sigma)$, does not decrease with longer training or larger model size, this result reflects a relative shift toward the training reference set rather than degraded similarity to validation data.

Together with the reconstruction-based analysis from \cref{sec:l2-metric}, these results show that the factors associated with a growing train–validation reconstruction gap—later checkpoints and larger models—are also associated with greater relative training affinity in trajectory-derived predictions. However, on the denoising training objective, they cause an eventual degradation in validation performance, while in inference, the absolute similarity to the validation set keeps increasing in all settings we tested.

\section{Strong conditioning amplifies memorization} \label{sec:strong-condition}
\begin{wrapfigure}{r}{0.5\textwidth}
    \vspace{-20pt}
    \centering
         \includegraphics[scale=0.6]{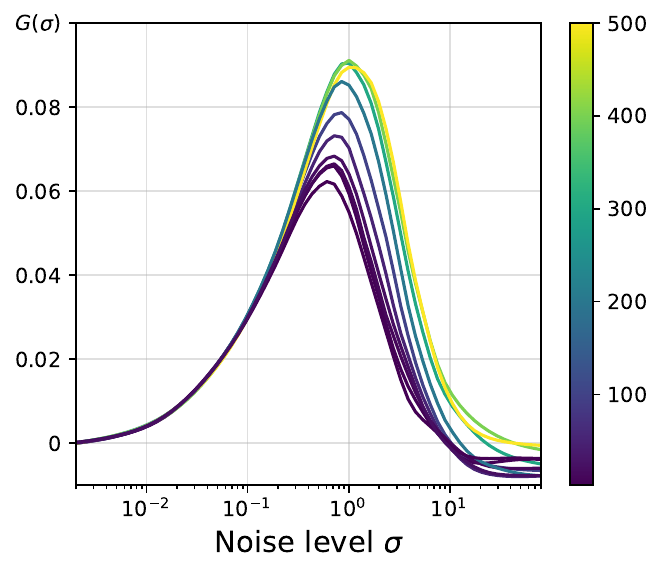}
   \caption{The relative generalization gap $G(\sigma)$ with average L2-reconstruction error increases and shifts towards higher noise levels when the number of classes (colorbar) for a class-conditioned model increases. Each line corresponds to a re-trained EDM model on CIFAR-10.}
    \label{fig:class-granularity}
    \vspace{-10pt}
\end{wrapfigure}

Because diffusion models operate in high-dimensional spaces, they contain many regions where the data manifold is sparsely supported. The data samples in these regions tend to be memorized and are not significantly less likely under the model than other data samples \citep{burg2021memorizationprobabilisticdeepgenerative}. 
This effect is exacerbated when modeling a conditional distribution, since it is supported by an even smaller subset of the dataset. Thus, model conditioning should amplify memorization, and the more granular the conditioning, the stronger the memorization, with text-to-image models being an extreme case. To test this, we retrained EDM \citep{karras2022edm} on CIFAR-10 for 100M images across varying numbers of classes. Classes were computed algorithmically using k-means clustering in DINOv2 feature space. 

Under the reconstruction-based metric from \cref{sec:l2-metric}, an increasingly granular model condition indeed exhibits an increased generalization gap (\cref{fig:class-granularity}a). This is consistent with prior work, which found that conditioning on even uninformative random labels triggers memorization in diffusion models \citep{gu2025memorizationdiffusionmodels}. Additionally, we find that this increase is more pronounced at higher noise levels and disappears entirely at lower noise levels, indicating that the condition primarily guides the early stages of the generation process. This is consistent with the toy model (\cref{fig:toy-gaps}b), though the effect here is weaker, likely because, unlike in the toy model, the conditional predictors are not independent across conditions. 
%Although the right-shift remains visible, the increase in the generalization gap appears to be mitigated when considering the L2-distance in feature space (\cref{fig:class-granularity}b). It is unclear whether this is a consequence of using features rather than images or whether the noise from applying a feature extractor to partially denoised samples obscures the trend. 

%Interestingly, FDD does not continue to improve indefinitely as the conditioning becomes more granular. Instead, it attains an optimal value for reasonably granular conditioning \citep{adaloglou2024rethinkingclusterconditioneddiffusionmodels}. It is unclear whether this is similar to the degradation observed at long training times, due to limitations of the metrics, or a separate mechanism inherent to model conditioning. 

\section{Conclusion, limitations and future work}
%In this work, we have shown that diffusion models generalize by two sources that prevent memorization: (i) the reconstruction error as measured by the training loss, which can be increased by reducing training times and model sizes, and (ii) the inability of denoising trajectories to follow the domain of noisy images seen during training. Whereas the generalisation gap of (i) is strongly increasing with training time and model size, the generalisation gap of (ii) remains remarkably stable.

In this work, we studied how structural memorization in diffusion models relates to both the denoising objective and the inference process. On the denoising objective, real diffusion models develop a growing train–validation reconstruction gap at intermediate noise levels that scales with longer training and larger model size, and is eventually accompanied by a decrease in validation performance. A fully analytic 2D toy model demonstrates how this gap arises from the interaction between model error and the density of the data distribution’s support: the optimal denoising flow field localizes sharply around training samples, whereas model error smooths the learned flow field and can suppress exact recall. Along inference trajectories, we find a related but distinct effect. Predictions made from intermediate states occupy a different feature-space regime from predictions made from directly noised training and validation images, yet their relative affinity shifts toward the training reference set as training progresses and model capacity increases. Unlike the reconstruction gap, this shift does not cause a decrease in absolute similarity to validation references.

The reconstruction gap and trajectory affinity are therefore not interchangeable measures, and the trajectory affinity alone does not establish a target-paired inference generalization gap. The former uses matched clean targets and directly measures denoising performance; the latter remains meaningful even when inputs have no matched targets. The inference result therefore complements the training-level picture: 
predictions from trajectory states do not resemble predictions from noised training images, but they do carry a bias towards the training set.

Our experiments are limited to class-conditional image generation using EDM and EDM2 architectures on ImageNet and CIFAR benchmarks. The reconstruction gap follows from general properties of the score-matching objective, but the quantitative form of the gap and the trajectory-affinity result may depend on the data modality, sampler, feature representation, and reference-set construction.

Beyond these methodological limitations, the results expose a measurement problem. Reconstruction-based metrics, verbatim-copy tests \citep{somepalli2023understandingmitigatingcopyingdiffusion,carlini2023extractingtrainingdatadiffusion,wen2024detectingexplainingmitigatingmemorization,hasegawa2025quantifyingeasereproducingtraining}, distributional distances, and trajectory-affinity measures probe different aspects of memorization and generalization, and no single metric captures the full phenomenon \citep{stein2023exposingflawsgenerativemodel}. A recent concurrent study likewise shows that standard test loss can coexist with increasing training-specificity in generated samples \citep{garnierbrun2026biasedgeneralizationdiffusionmodels}, complementing our analysis. An important open question is how these relative affinity shifts relate to final-sample copying, privacy leakage, and downstream sample quality, and whether they can be altered independently through training or inference interventions.

\subsubsection*{Author Contributions}
Tim Kaiser contributed all code and experiments, the literature research for the manuscript, the structure of the paper and its narrative, the writing of the manuscript and its final polish.
Prof. Dr. Markus Kollmann contributed the initial idea of the relative generalization gap with average L2-reconstruction error, as well as guidance and discussion of the results, narrative, and manuscript.

% \bibliography{iclr_files/iclr2027_conference}
\bibliography{main}
\bibliographystyle{iclr_files/iclr2027_conference}
\clearpage 
\appendix
\begin{center}
  {\Large Beyond Sample Copying: Structural Memorization in Diffusion Models}

  {\large \textbf{Supplementary Material}}
\end{center}
\vspace{2mm}
This document provides details and additional experiments supporting the main text. In \cref{supp:relatedwork}, we review the related literature in detail. In \cref{supp:reproducibility}, we provide the necessary details to reproduce our results. In \cref{supp:guidance}, we investigate how the relative generalization gap behaves when using diffusion guidance. In \cref{supp:longrange}, we provide additional experiments about the impact of the modelling of long- and short-range dependencies on memorization. In section \cref{supp:toy-details}, we report results for the toy model with random instead of symmetric data. In \cref{supp:additional-gengaps} and \cref{supp:additional-knn}, we provide additional results supporting the claims we made in \cref{sec:l2-metric} and \cref{sec:inference}.

\section{Related work} \label{supp:relatedwork}
\paragraph{Background.} 
Diffusion models were developed in their current form from two different approaches: via variational inference with an objective resembling denoising score matching \citep{ho2020denoising}, and by time-reversed stochastic differential equations with direct application of score matching \citep{song2020generativemodelingestimatinggradients, song2021scorebased}. Several improvements, including architectural advances \citep{dhariwal2021diffusionmodelsbeatgans}, sampling techniques such as guidance \citep{dhariwal2021diffusionmodelsbeatgans}, and latent diffusion \citep{rombach2022highresolutionimagesynthesislatent}, have established diffusion models as the dominant class of generative models. This work uses the EDM framework developed by Karas et al \citep{karras2022edm}, which achieved state-of-the-art performance in the class-conditional setting, owing to a comprehensive suite of analyses and improvements to the training and sampling of diffusion models \citep{karras2022edm, karras2024edmv2}. %This work uses the EDM Diffusion formulation, where the targets used in the forward pass are real data samples, as opposed to the formulation that uses a linear interpolation between data samples and trajectory starting points as targets, known as the Rectified Flow formulation or optimal transport paths \citep{liu2022flowstraightfastlearning}. 
% Early work on \textbf{memorization and generalization in probabilistic generative models} established \tim{how memorization in unsupervised density estimation differs from memorization in supervised learning} \citep{burg2021memorizationprobabilisticdeepgenerative}. 

\paragraph{Understanding generalization in diffusion models.} 
Recently, analytical solutions for denoising trajectories have been established for simple or small-sized datasets, showing that optimal diffusion models are unable to sample anything but the training data \citep{scarvelis2025closedformdiffusionmodels,adaloglou2025swg, gao2024flowmatchingmodelsmemorize}. %This is in stark contrast to the classical view on memorization, where a fully memorizing model can technically still generalize perfectly. Following this, %and in light of the manifold hypothesis \citep{fefferman2013testingmanifoldhypothesis}, 
% a series of works investigated the relationship between optimal diffusion and well-generalizing real diffusion models. 
This is in stark contrast to the impressive ability of real diffusion models to generate high-quality, novel samples across different data modalities. Tackling this paradox, a series of works showed that the optimal flow field has a sharp geometry around the training samples \citep{gao2024flowmatchingmodelsmemorize, ventura2025manifoldsrandommatricesspectral} and that smoothing this geometry plays a key role for successful generalization \citep{farghly2025diffusionmodelsmanifoldhypothesis, achilli2024losingdimensionsgeometricmemorization, scarvelis2025closedformdiffusionmodels}. Several works have further characterized this smoothing through different lenses: as convergence toward the Gaussian structure of the training data \citep{li2024understandinggeneralizabilitydiffusionmodels}, as an effect of variance in the score matching target \citep{vastola2025generalizationvariancenoiseshapes}, as an effect of using a neural network to approximate the score \citep{zhou2026smoothingscorefunctiongeneralization}, and as a regularization problem \citep{baptista2025memorizationregularizationgenerativediffusion}. Possible causes of this smoothing include locality \citep{lukoianov2025localityimagediffusionmodels} and equivariance \citep{kamb2025analytictheorycreativityconvolutional} biases from both data statistics and the architecture, a bias towards geometry-adaptive harmonic bases adapted to the underlying images \citep{kadkhodaie2024generalizationdiffusionmodelsarises}, and implicit regularization \citep{farghly2025implicitregularisationdiffusionmodels}. Finally, it has been shown that a smoothed flow field causes diffusion models to interpolate between modes of the training data \citep{chen2026interpolationeffectscoresmoothing,aithal2024understandinghallucinationsdiffusionmodels}. Training dynamics further modulate this: training time \citep{bonnaire2025diffusionmodelsdontmemorize}, dataset size \citep{halder2025solvablegenerativemodellinear, favero2025biggerisntmemorizingearly}, and model complexity all affect when and how strongly memorization appears, alongside less obvious factors such as differences in training loss \citep{buchanan2025edgememorizationdiffusionmodels}, the number of noise samples per data point \citep{george2025denoisingscorematchingrandom}, and model conditioning \citep{gu2025memorizationdiffusionmodels}.

\paragraph{Detecting memorization in diffusion models.} 
Prior work on detecting memorization has largely focused on sample-level detection at inference: Nearest-neighbor and copy-detection approaches \citep{somepalli2022diffusionartdigitalforgery,somepalli2023understandingmitigatingcopyingdiffusion,kowalczuk2025findingdorimemorizationtexttoimage,dombrowski2025lcmemuniversalmodelrobust} flag generated images that closely match a specific training sample. Extraction attacks \citep{carlini2023extractingtrainingdatadiffusion,wen2024detectingexplainingmitigatingmemorization} recover memorized images by adversarially probing the model. Model-centric scores such as Effective Model Memorization \citep{gu2025memorizationdiffusionmodels} and ODE-volume-based quantification \citep{hasegawa2025quantifyingeasereproducingtraining} summarize a model's overall tendency toward verbatim reproduction.

\paragraph{Training-bias beyond standard generalization metrics.}
A recent line of work has questioned whether decreasing held-out loss is sufficient to characterize generalization in diffusion models. Garnier-Brun et al. identify a phase of biased generalization in which test loss continues to decrease while generated samples acquire anomalously high proximity to the training data, and attribute this effect to the progressive learning of finer, more training-dependent features \citep{garnierbrun2026biasedgeneralizationdiffusionmodels}. Their analysis operates primarily at the level of final generated samples and their proximity to the training set. We study the relationship between this kind of training-specificity and the denoising process itself, at all noise levels, and in three regimes of the input space: Noisy training images, noisy held-out validation images, and the intermediate states of inference trajectories. The resulting trajectory measure is complementary to, rather than a substitute for, final-sample proximity or held-out loss.

\paragraph{The role of the noise level.}
The noise level at which memorization manifests has received particular attention. Statistical-physics analyses of the optimal score predict a "collapse" regime at small noise, where trajectories are pulled into individual training points \citep{Biroli_2024, ventura2025manifoldsrandommatricesspectral}. Closely related to our analysis of the relative generalization gap, a complementary geometric account argues that memorization risk is highly non-uniform across noise levels, and identifies a danger zone at intermediate noise levels where memorization is most pronounced \citep{dodson2026calmendswildmiddle}. Our analysis agrees that the reconstruction gap is concentrated at intermediate noise, but asks a different question at inference: how do predictions from trajectory states relate to training and validation references? We find that they develop greater relative affinity to training references as training progresses and model capacity increases, without a corresponding decrease in their absolute similarity to validation references.

\paragraph{The role of the denoising objective}
% The denoising training objective is prone to memorization by construction, as the aforementioned optimal diffusion models are its unique global optimum \citep{gu2025memorizationdiffusionmodels}. 
Closely related to our finding about the interplay between sampling trajectories and the likely locations of noisy training images in input space, Song et al. refine the perspective of a globally smoothed score function \citep{song2025selectiveunderfittingdiffusionmodels}. They find that better diffusion models selectively fit the score in some regions of the input space, while underfitting it in others. As models improve, their score approximation improves in the regions of input space the model is highly likely to see during training, while it worsens outside of those regions. As we do, they observe that inference trajectories and noisy training images have distinct geometries in feature space. Unlike Song et al., we quantify this asymmetry at the reconstruction level and its expression along inference trajectories.

% \textbf{The role of metrics.} Memorization and overfitting are of growing concern for the performance-oriented literature on diffusion models, where they have largely been absent from the discussion. As quantitative phenomena, they depend on the metric that is used to measure them. This proves to be a difficult task, as established generative metrics focus on measuring fidelity and diversity, not novelty \citep{jiralerspong2024featurelikelihooddivergenceevaluating}, and memorization-specific metrics are unable to distinguish it from overfitting \citep{theis2016noteevaluationgenerativemodels, stein2023exposingflawsgenerativemodel}. %\new{Our work differs from the above in that we study a continuous, structural form of memorization — the denoiser's differential treatment of training and unseen data — which, in general, does not manifest as verbatim reproduction.}

\section{Reproducibility} \label{supp:reproducibility}
\subsection{Technical details}
\begin{itemize}[nosep]
  \item \textbf{Feature extractors.} We used the official implementation from \citep{karras2024guidingdiffusionmodelbad} for the feature detectors DINOv2 ViT-L/14 and Inception-V3.
  \item \textbf{Noise schedule.} We used the official hyperparameters from \citep{karras2024edmv2}, i.e., $\sigma \in [2e-3,80]$. 
  \item \textbf{Sampler.} We used the official EDM/EDM2-sampler from \citep{karras2024edmv2} with default hyperparameters.
  \item \textbf{EDM preconditioning.} For all experiments, we use the prediction after preconditioning, i.e., the full target prediction. Any other choice would lead to distortions based on the preconditioning parameters.
\end{itemize}

\subsection{Average reconstruction error details} \label{supp:l2-details}
\paragraph{Input construction} We construct all denoiser inputs by directly noising a data point $\y \in {\cal D}_\text{train/val}$: $\x(\sigma) = \y + \sigma \eta$. We match the noise realization $\eta \sim \mathcal{N}(0,I)$ to each image $\y$ via its random seed, so it remains fixed across noise levels $\sigma$, model sizes, and model checkpoints. We sweep 64 noise levels on a log scale within $\sigma \in [2e-3, 80]$, with 8192 samples each.
\begin{algorithm}[h!]
\caption{$L({\cal D}, \sigma)$ (\cref{eq:l2-error}}
\label{alg:l2}
\begin{algorithmic}[1]
  \Require Denoiser $D$, reference set with class-labels ${\cal D}=\{(\y_i,c_i)\}$, noise level $\sigma$
  \For{$\y_i \in \mathcal{R}$}
    \State $\x_i(\sigma) \gets \y_i + \sigma \eta, \eta \sim N(0,I)$
    \State Compute one-step denoiser prediction $\hat{\y}_i \gets D(\x_i(\sigma),\,\sigma,\,c_i)$
  \EndFor
  \State $L({\cal D}, \sigma) \gets \frac{1}{N} \sum_i ||\hat{\y}_i - \y_i||_2^2$
  \State \Return $L({\cal D}, \sigma)$
\end{algorithmic}
\end{algorithm}

\subsection{Average cosine similarity details} \label{supp:knn-details}
\paragraph{Query construction.}
For both metrics, each query is first passed through the denoiser at the
specified noise level. Let $\phi(\cdot)$ denote the feature extractor and
let $\hat{\q}_i$ denote the resulting one-step prediction. For directly
noised queries, the denoiser input is
$\x_i(\sigma)=\q_i+\sigma\eta_i$. For trajectory queries, $\x_i(\sigma)$
is the saved diffusion trajectory state at the corresponding noise level.
The denoiser and feature-extraction steps are therefore identical across
query populations; only the construction of the denoiser input differs.

\paragraph{kNN feature similarity.}
For each query prediction $\hat{\q}_i$, we compute cosine similarities to
reference features from the same class,
\[
 a_{ij}=\cos\!\left(\phi(\hat{\q}_i),\phi(\r_j)\right),
 \qquad
 \mathcal{J}_i=\{j:c_j=c_i\}.
\]
Let $\mathcal{N}_k(i;R)$ denote the indices of the $k$ largest values of
$a_{ij}$ among $j\in\mathcal{J}_i$. The kNN similarity is
\[
 S_k(Q,R;\sigma)
 =\frac{1}{|Q|}\sum_{i=1}^{|Q|}
 \frac{1}{k}\sum_{j\in\mathcal{N}_k(i;R)}a_{ij}.
\]
Thus, $S_k(Q,R;\sigma)$ is the mean cosine similarity between each query
prediction and its $k$ most similar same-class reference images, averaged
first over neighbors and then over queries. The reference features are
computed once and reused across query populations, noise levels, checkpoints,
and model sizes.

For the train--validation affinity analysis, we evaluate each query
population against both reference sets and define
\[
 B_k(Q;\sigma)
 =S_k(Q,R_{\mathrm{train}};\sigma)
 -S_k(Q,R_{\mathrm{val}};\sigma).
\]
Positive values indicate greater feature-space affinity to the training
reference set than to the validation reference set. Exact self-matches are
excluded from $S_k(Q,R_{\mathrm{train}})$ for training queries and from
$S_k(Q,R_{\mathrm{val}})$ for validation queries.

\begin{algorithm}[h!]
\caption{$S_k(Q,R;\sigma)$ (\cref{eq:similarity}}
\label{alg:sq}
\begin{algorithmic}[1]
  \Require Denoiser $D$, query set $Q=\{(\q_i,c_i)\}_{i=1}^{N_Q}$,
  reference set $R=\{(\r_j,c_j)\}_{j=1}^{N_R}$, noise level $\sigma$,
  feature extractor $\phi$, number of neighbors $k$
  \For{$j=1,\ldots,N_R$}
    \State $v_j\gets\phi(\r_j)$
  \EndFor
  \For{$i=1,\ldots,N_Q$}
    \State Construct the denoiser input $\x_i(\sigma)$
    \State $\hat{\q}_i\gets D(\x_i(\sigma),\sigma,c_i)$
    \State $\bm u_i\gets\phi(\hat{\q}_i)$
    \State $\mathcal{J}_i\gets\{j:c_j=c_i\}$
    \State $a_{ij}\gets\cos(\bm u_i,\bm v_j)$ for all $j\in\mathcal{J}_i$
    \State Select $\mathcal{N}_k(i;R)$, the indices of the $k$ largest $a_{ij}$
    \State $s_i\gets\frac{1}{k}\sum_{j\in\mathcal{N}_k(i;R)}a_{ij}$
  \EndFor
  \State $S_k(Q,R;\sigma)\gets\frac{1}{N_Q}\sum_{i=1}^{N_Q}s_i$
  \State \Return $S_k(Q,R;\sigma)$
\end{algorithmic}
\end{algorithm}

\subsection{Compute resources} \label{supp:compute}
Below is a table detailing the compute resources required to run the experiments of this paper. We used 4 NVIDIA A100 GPUs (45GB) for all experiments. FDD and FID are computed in parallel, as they share everything but their feature extractor. Full experiments consist of runs across many model sizes, checkpoints, noise levels, and possibly additional parameters, such as guidance weights. Due to the large number of forward passes per sample, standard FD computation is highly sensitive to model size. In contrast, the reconstruction-based metrics are much less sensitive, since there, feature extraction, FD computation, and decoding share a relatively larger part of the total cost.
\begin{table}[h]
  \centering
  \caption{Compute requirements for metric evaluation on ImageNet-64, for one model (checkpoint) and a single noise level $\sigma$. GPU-h and wall-clock time for ImageNet range from smallest to largest model size. They are approximate and refer to a server with 4xNVIDIA-A100 (45GB).}
  \label{tab:compute}
  \smallskip
  \begin{tabular}{lllrr}
    \toprule
    \textbf{Dataset} & \textbf{Metric} & \textbf{Sample size} & \textbf{GPU-h} & \textbf{Wall-clock} \\
    \midrule
    ImageNet-64 & $L({\cal D}, \sigma)$ & 8192 & 4m - 4m & 1m - 1m \\
    % ImageNet-64 & FD with $\x_t(\sigma)$ & 8192 & &  \\
     & Standard FD & 50.000 & 56m - 4h20m & 14m - 1h5m \\
    & $S_k(Q, R; \sigma)$ & 10.000 & 4m - 4m & 1m - 1m \\
    \midrule
    ImageNet-512 & $L({\cal D}, \sigma)$  & 8192 &  16m - 16m & 4m - 4m \\
    % ImageNet-512 & FD with $\x_t(\sigma)$ & 8192 & &   \\
     & Standard FD & 50.000 & 2h28m - 7h44m & 37m - 1h56m \\
    & $S_k(Q, R; \sigma)$ & 10.000 & - & -\\
    \midrule
    CIFAR-10/100 & $L({\cal D}, \sigma)$  & 8192 & 4m & 1m \\
    % CIFAR-10/100 & FD with $\x_t(\sigma)$ & 8192 & & \\
     & Standard FD & 50.000 & 8m & 2m \\
    & $S_k(Q, R; \sigma)$ & 10.000 & 8m & 2m\\
    \bottomrule
  \end{tabular}
\end{table}

\section{Diffusion guidance} \label{supp:guidance} 
% \begin{wrapfigure}{r}{0.5\textwidth}
%     \vspace{-10pt}
%     \centering
%     \includegraphics[width=0.48\textwidth]{imgs/gap-in64-edm2-s-auto-dino.pdf}
%     \caption{Guidance improves sample quality and distribution fit without compromising generalization. Generalization gaps $(M_{val} - M_{train})/M_{train}$ for Autoguidance with EDM2-S on ImageNet-64 with varying guidance strengths. Metric A-C fix $\sigma$ at the peak in the generalization gap.}
%     \label{fig:gen-gap-guidance}
%     \vspace{-20pt}
% \end{wrapfigure}
\begin{figure}[h]
     \centering
     \begin{subfigure}[t]{0.32\textwidth}
         \centering
         \includegraphics[scale=0.4]{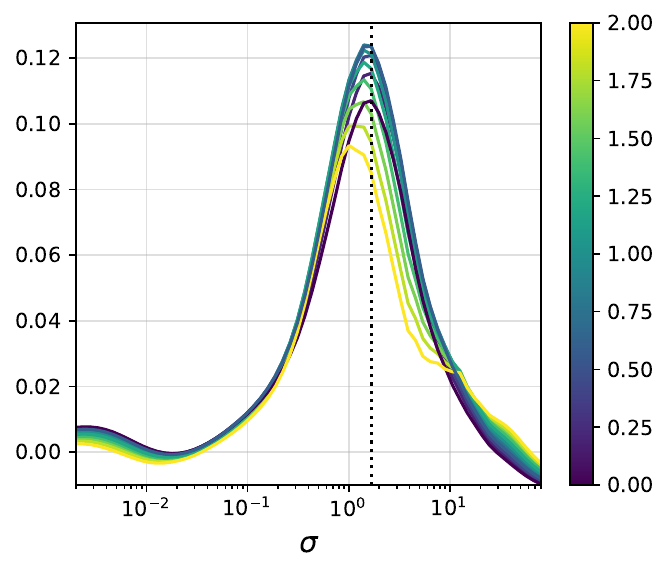}
         \caption{CFG}
     \end{subfigure}
     \begin{subfigure}[t]{0.32\textwidth}
         \centering
         \includegraphics[scale=0.4]{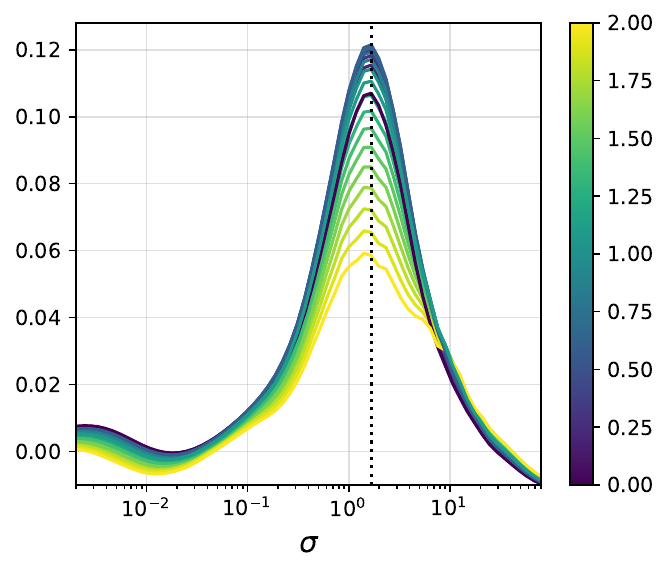}
         \caption{Autoguidance}
     \end{subfigure}
     \begin{subfigure}[t]{0.32\textwidth}
         \centering
         \includegraphics[scale=0.4]{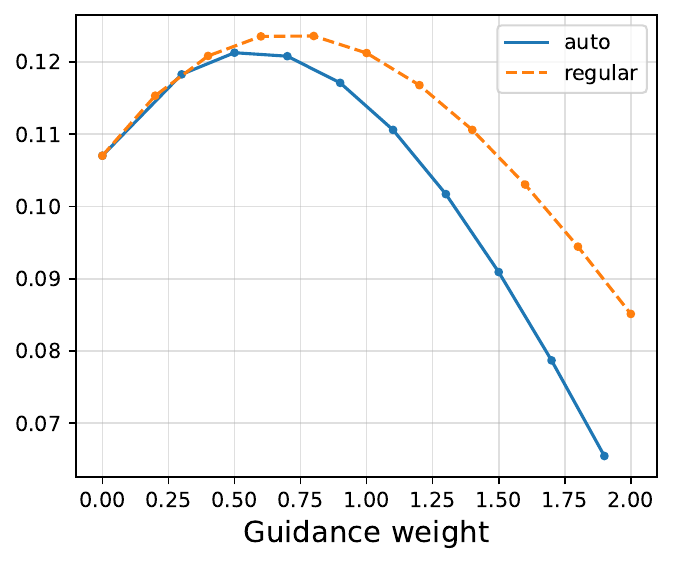}
         \caption{$\sigma \approx 1.67$}
     \end{subfigure} 
     \caption{Relative generalization gaps (\cref{eq:gen-gap}) with EDM2-S on ImageNet-64. $w=0$ corresponds to no guidance. An increasing guidance weight amplifies the gap, but this trend reverses for moderate to high guidance weights. (c) fixes $\sigma$ close to the peak of the generalization gaps in (a) and (b), indicated by the dotted black line.}
    \label{fig:gen-gap-guidance}
\end{figure}
Diffusion guidance is a sampling technique that enhances sample quality by correcting biases in the diffusion model using an auxiliary diffusion model that exhibits these biases even more strongly \citep{adaloglou2025swg}. Classifier-Free Guidance (CFG) \citep{ho2022classifierfreediffusionguidance} uses an unconditional model to amplify conditioning, whereas other methods use auxiliary models to correct or accentuate particular biases \citep{ahn2025selfrectifyingdiffusionsamplingperturbedattention,adaloglou2025swg,hong2024smoothedenergyguidanceguiding}. The best results (on FDD) are obtained when the auxiliary model is a worse version of the primary model, e.g., by having fewer parameters or less training time (Autoguidance \citep{karras2024guidingdiffusionmodelbad}). In our toy model, this corresponds to combining predictors with different levels of model error. The resulting guided predictor more closely approximates the optimal denoiser and thereby sharpens the learned flow field \citep{adaloglou2025swg}. Since an increased condition granularity amplifies the generalization gap (\cref{sec:strong-condition}), the same might hold for CFG, because it accentuates the model condition. The toy model predicts a moderate yet steady increase in the generalization gap as the guidance weight increases, given that the primary and auxiliary models differ only in their model error $\delta$ (\cref{fig:toy-gaps}c). However, at larger guidance weights, guidance amplifies small differences between the models too much, like color saturation, which degrades overall performance. This is not reflected in the toy model, due to its simplicity. %It is known that, for moderate guidance weights, Autoguidance corrects the prediction toward the prediction of the optimal predictor \citep{adaloglou2025swg}, thereby sharpening the flow field around training points, which should result in a larger generalization gap. 

In practice, we see exactly these two trends play out: For small to moderate guidance weights, the gap increases, whereas larger guidance weights significantly reduce it (\cref{fig:gen-gap-guidance}). This trend is similar for both CFG and Autoguidance. Note that optimal performance, as measured by FDD, is achieved around $w = 0.9$ for CFG and $w=1.4$ for Autoguidance. When increasing the weight further, the degradation due to the overcorrection of small differences between the models outweighs the improvements due to the correction of the right biases \cite{adaloglou2025swg}.

%Surprisingly, in practice, although \md{} behaves similarly for increasing guidance weight as for increasing training time, the reconstruction-based metrics do not, neither for Autoguidance nor for CFG (\cref{fig:gen-gap-guidance}). Not only are the generalization gaps for these metrics insensitive to the guidance weight, but the performances on the training and validation splits are also flat (\cref{fig:gen-gap-guidance}c). %This suggests that there are ways to improve sample quality and the fit of the distribution without compromising the model's generalization abilities.
%This indicates that diffusion guidance has very little effect on the model's reconstruction performance, despite its significant benefits for regular inference. 

\section{Long-range dependencies tend to be memorized} \label{supp:longrange}
\begin{figure}[bh]
     \centering
     \begin{subfigure}[b]{0.32\textwidth}
         \centering
         \includegraphics[scale=0.34]{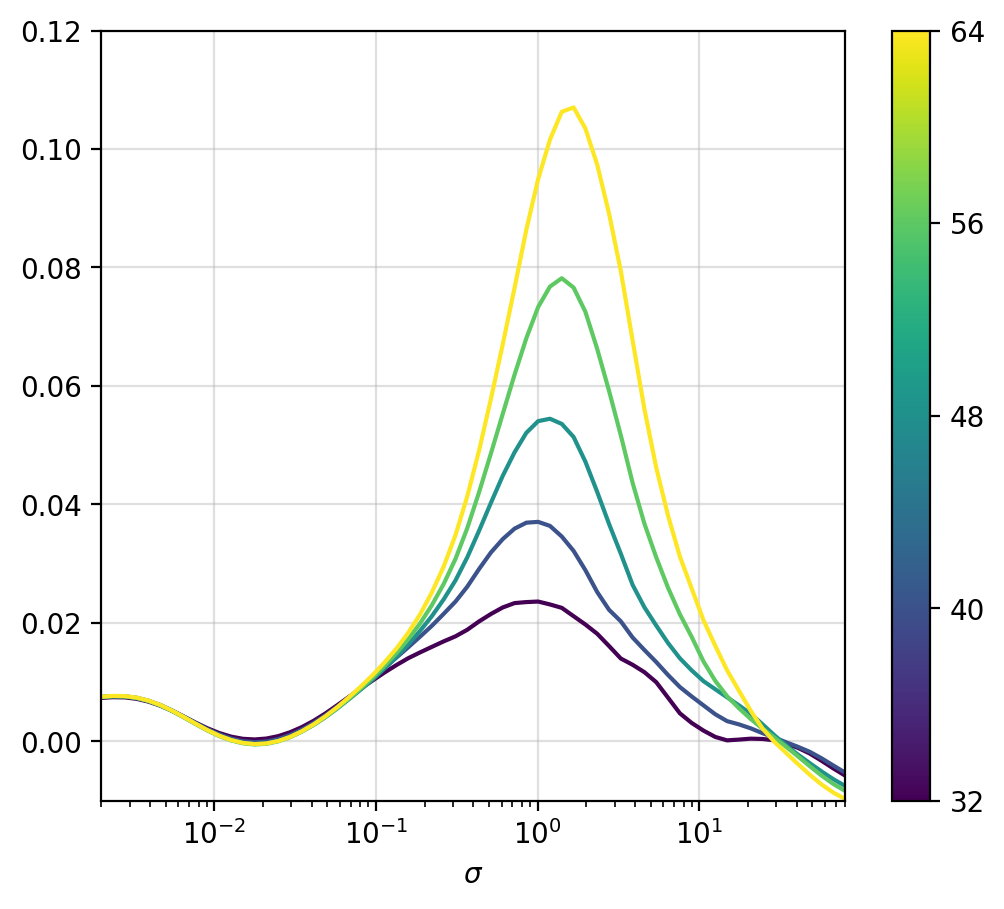}
         \caption{Reduced receptive field}
     \end{subfigure}
     \begin{subfigure}[b]{0.28\textwidth}
         \centering
         \includegraphics[scale=0.34]{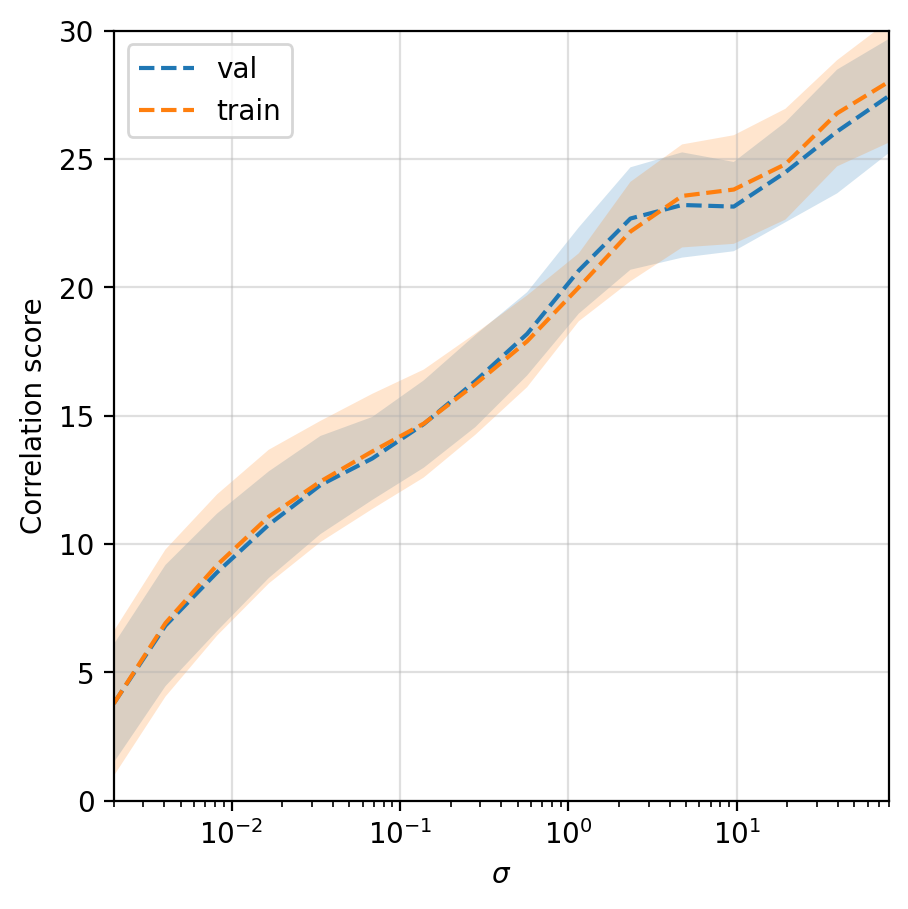}
         \caption{Avg. correlation score}
     \end{subfigure}
     \begin{subfigure}[b]{0.38\textwidth}
         \centering
         % Nudge this plot UP by 5mm to compensate for the title
        \raisebox{3mm}{\includegraphics[scale=0.4]{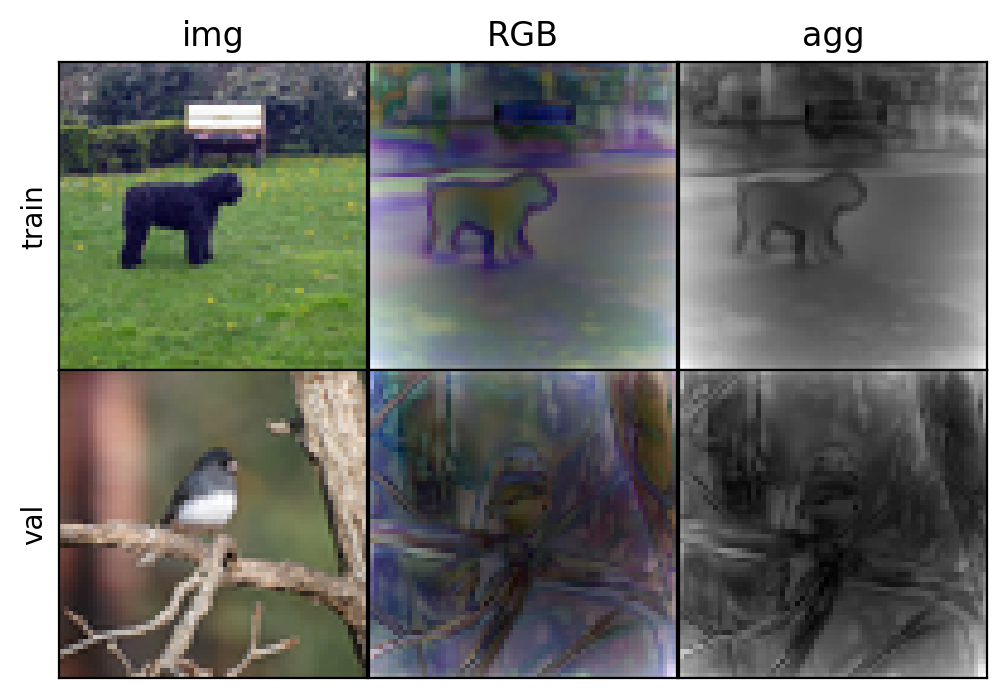}}
        \caption{Pixel-wise correlation score}
     \end{subfigure}
     \caption{\textbf{(a)} Decreasing the receptive field mitigates the relative generalization gap (\cref{eq:gen-gap}) with $M_\text{L2}$. Colorbar shows the sliding window size. 64 corresponds to regular inference. \textbf{(b)} During inference, the model shifts from long-range correlations to short-range correlations. The correlation score is averaged over pixels and images. \textbf{(c)} Edge pixels tend to have longer-range dependencies. $\sigma \approx 1.2$. RGB shows the pixel-wise correlation score, agg is aggregated over channels. All plots show ImageNet-64, EDM2-S.}
     \label{fig:swg-corr-object}
\end{figure}
% \begin{figure}[!h]
%      \centering
%      \begin{subfigure}[b]{0.65\textwidth}
%          \centering
%          \includegraphics[scale=0.55]{imgs/edm2-gradients_CS_both_26.png}  
%          % \includegraphics[scale=0.45]{imgs/edm2-gradients_CS_both_26_full.png}
%          \caption{Single image}
%      \end{subfigure} 
%      \hfill
%      \begin{subfigure}[b]{0.34\textwidth}
%          \centering
%          \includegraphics[scale=0.55]{imgs/edm2-gradients_CS_both_avg.png}
%          \caption{128 images}
%      \end{subfigure}
%      \caption{Edge pixels tend to have longer-range dependencies. RGB shows the channel-wise correlation score, agg the mean over channels. \textbf{(a)} For a single image. \textbf{(b)} Averaged over 128 images. EDM2-S on ImageNet-512 at $\sigma=1$.}
%      \label{fig:gradients}
% \end{figure}
\begin{wrapfigure}{r}{0.3\textwidth}
    \vspace{-10pt}
    \centering
    \includegraphics[width=0.2\textwidth]{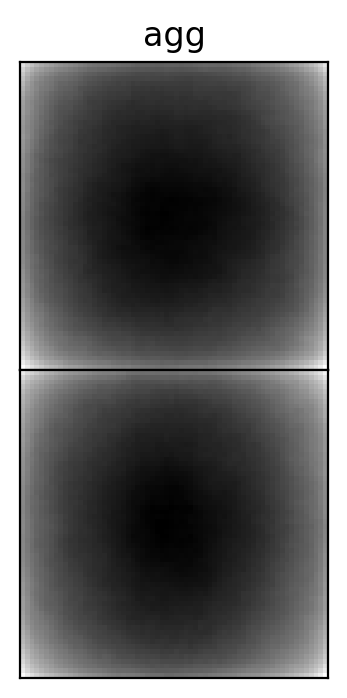}
    \caption{Edge pixels tend to have longer-range dependencies. Shown is the pixel-wise correlation score, averaged over 128 images and aggregated over channels, computed with EDM2-S on ImageNet-64 at $\sigma=1$.}
    \label{fig:cs-score-avg}
    \vspace{-10pt}
\end{wrapfigure}
Globally coherent images are more difficult to generate because features 
that require correct long-range pixel dependencies are highly image-specific and thus harder to generalize than local pixel patterns. Similar to training samples in sparsely supported regions of the input space, this should promote memorization of said features. Similar hypotheses were formulated by Niedoba et al.\ \citep{niedoba2025mechanisticexplanationdiffusionmodel}, who showed that aggregating local empirical denoisers replicates network generalization behavior, and An et al.\, who showed that restricting attention locality in diffusion transformers improves generalization \citep{an2024inductivebiasesenablegeneralization}. To test this, we constrain the range of pixel dependencies the denoiser can model, following \citep{adaloglou2025swg}, by denoising smaller image patches of equal size and then composing them back together. Overlapping pixels are averaged. 
We find that this dramatically shrinks the generalization gap (\cref{fig:swg-corr-object}a), supporting the hypothesis. 

Further, we investigate how the denoising of each pixel is affected by other pixels, how these dependencies are distributed within an image, and how this relates to the noise level $\sigma$(t).
%Additionally, when investigating how the denoising of pixels is affected by other pixels, we find that models are most focused on long-range dependencies at the beginning of inference, with the distance of the average gradient signal across an image strictly decreasing with $\sigma(t)$ (\cref{fig:corr-swg}). 
We define a pixel-level correlation score by computing the average gradient of a pixel in the denoised image with respect to all pixels in the input image, weighted by the L2 distance to those pixels. Specifically, given an image $\y$ with $N$ pixels, a denoiser $D(.)$, noise level $\sigma$ and noise $\n \sim \mathcal{N}(0,1)$, we compute the correlation score of the $ij$-th pixel as 
\[
CS_{ij} = \frac{\frac{1}{N}\sum_{k, l} d(i,j,k, l)\frac{\partial D(\x)_{ij}}{\partial \y_{kl}}}{\frac{1}{N} \sum_{k, l} \frac{\partial D(\x)_{ij}}{\partial \y_{kl}}} , \quad \x = \y + \sigma \n, \n \sim \mathcal{N}(0,I)
\]
with $d(i,j,k, l) = \bigg\Vert\begin{pmatrix}i \\ j\end{pmatrix} - \begin{pmatrix}k \\ l\end{pmatrix}\bigg\Vert_2$ being the L2 distance between the $ij$-th and $kl$-th pixels.
%Interestingly, edge pixels tend to take information from the center but not vice versa, i.e., the object determines the background, but the background does not determine the object (\cref{fig:gradients}). \tim{[Check this by investigating them before aggregating with the distance matrix.]}
A larger correlation score for a given pixel indicates that the gradient activations are high for distant pixels. 
We find that, on average, the farther a pixel in the output image is from the center, the longer its dependencies extend across the input images (\cref{fig:swg-corr-object}c, \cref{fig:cs-score-avg}). In other words, the object determines the background, but the background does not determine the object. It is unclear how the dataset's object-centric nature influences this result. When averaging the correlation scores across images, we see that as the noise level $\sigma(t)$ decreases, the distance of the average gradient signal across an image also decreases (\cref{fig:swg-corr-object}b), i.e., the focus shifts from long-range dependencies to short-range dependencies during inference. 

\section{Toy model with random data} \label{supp:toy-details}
\begin{figure}[h!]
     \centering
     \begin{subfigure}[b]{0.32\textwidth}
         \centering
         \includegraphics[scale=0.55]{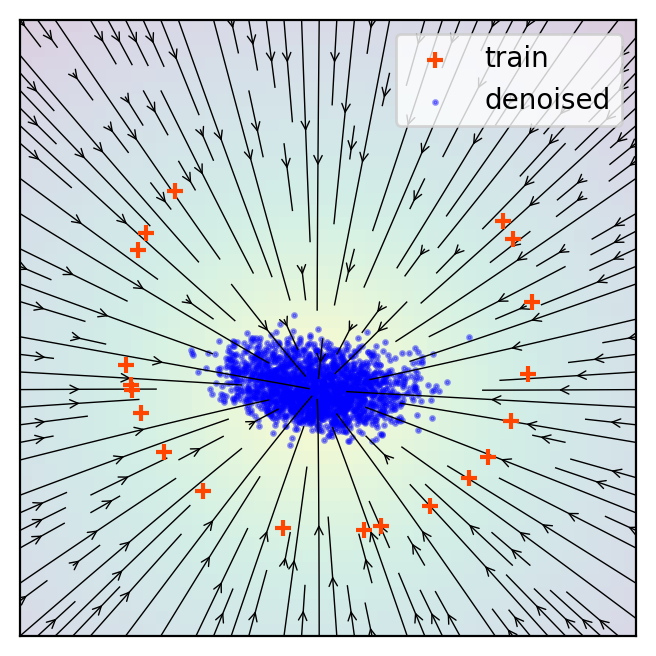}
         \caption{High noise}
     \end{subfigure} 
     \begin{subfigure}[b]{0.32\textwidth}
         \centering
         \includegraphics[scale=0.55]{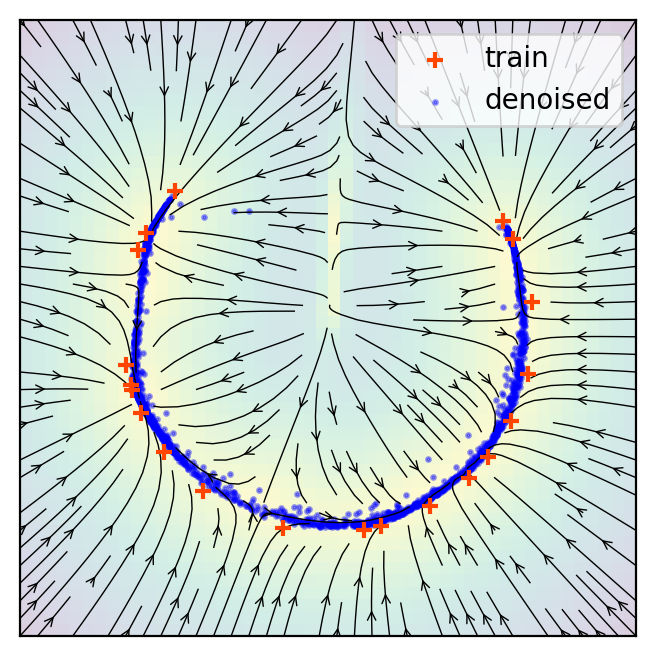}
         \caption{Intermediate noise}
     \end{subfigure}
     \begin{subfigure}[b]{0.32\textwidth}
         \centering
         \includegraphics[scale=0.55]{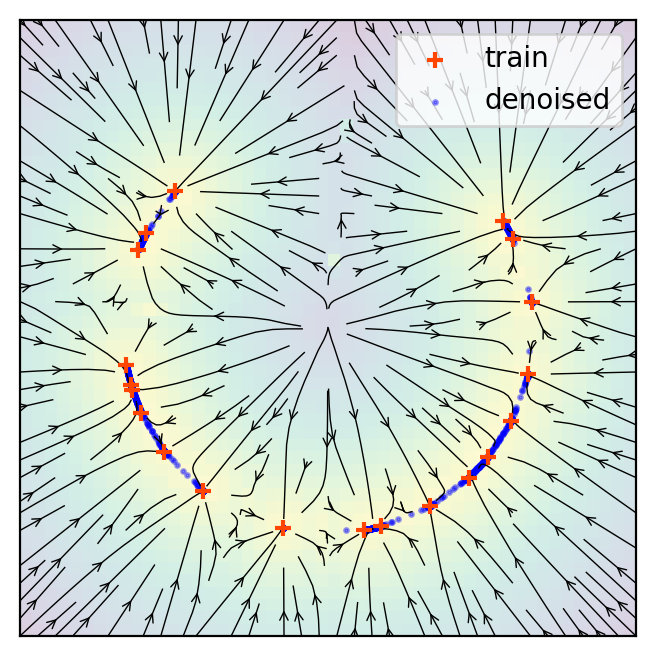}
         \caption{Low noise}
     \end{subfigure}
     \caption{Flow field lines and predictions of the optimal target predictor $\y^*(\x, \sigma)$ at different noise levels $\sigma = 28, 2.8, 0.63$. Color indicates the magnitude of the prediction error $\y^*(\x, \sigma) - x$. Field geometry is \textbf{(a)} global, predictions tend towards the superposition of all training points. \textbf{(b)} moderately localized around training points, predictions approximate the data manifold \textbf{(c)} highly localized around each training point, predictions replicate the training points.}
     \label{fig:toy-geometry-rnd}
\end{figure}

\begin{figure}[h]
     \centering
     \begin{subfigure}[b]{0.32\textwidth}
         \centering
         \includegraphics[scale=0.35]{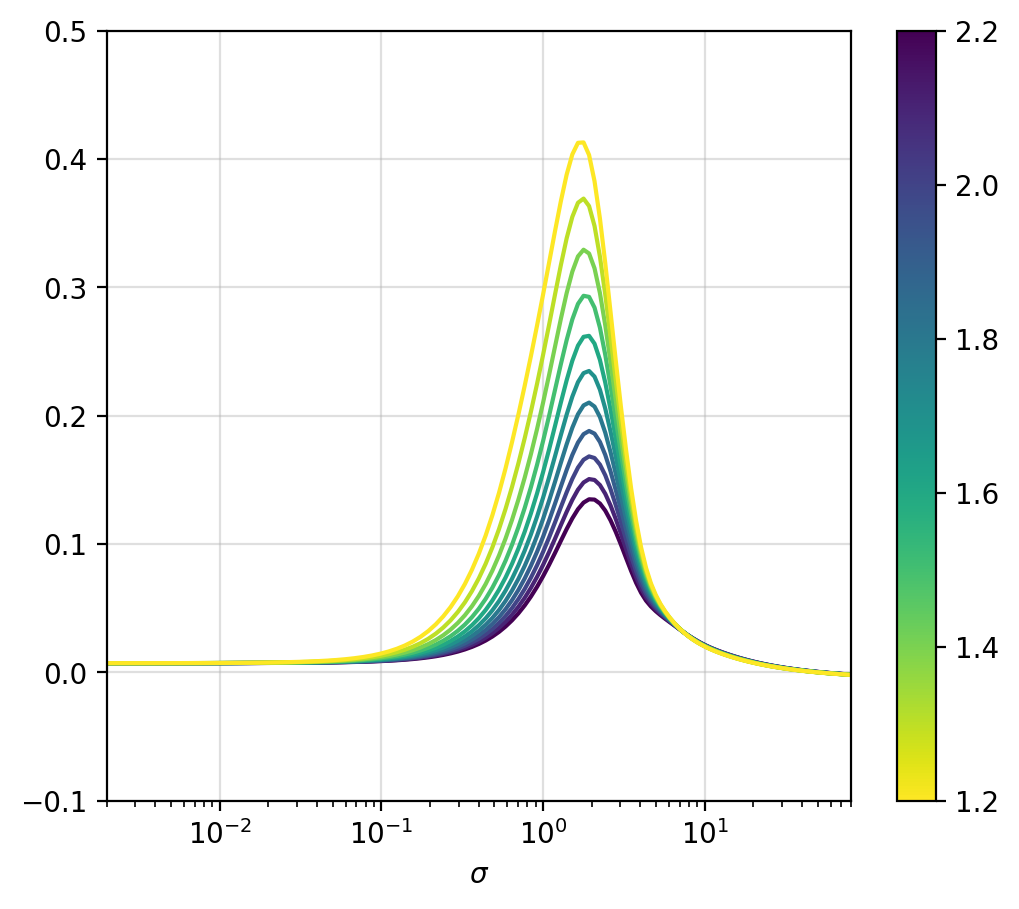}
         \caption{Error parameter $\delta$}
     \end{subfigure} 
     \begin{subfigure}[b]{0.32\textwidth}
         \centering
         \includegraphics[scale=0.35]{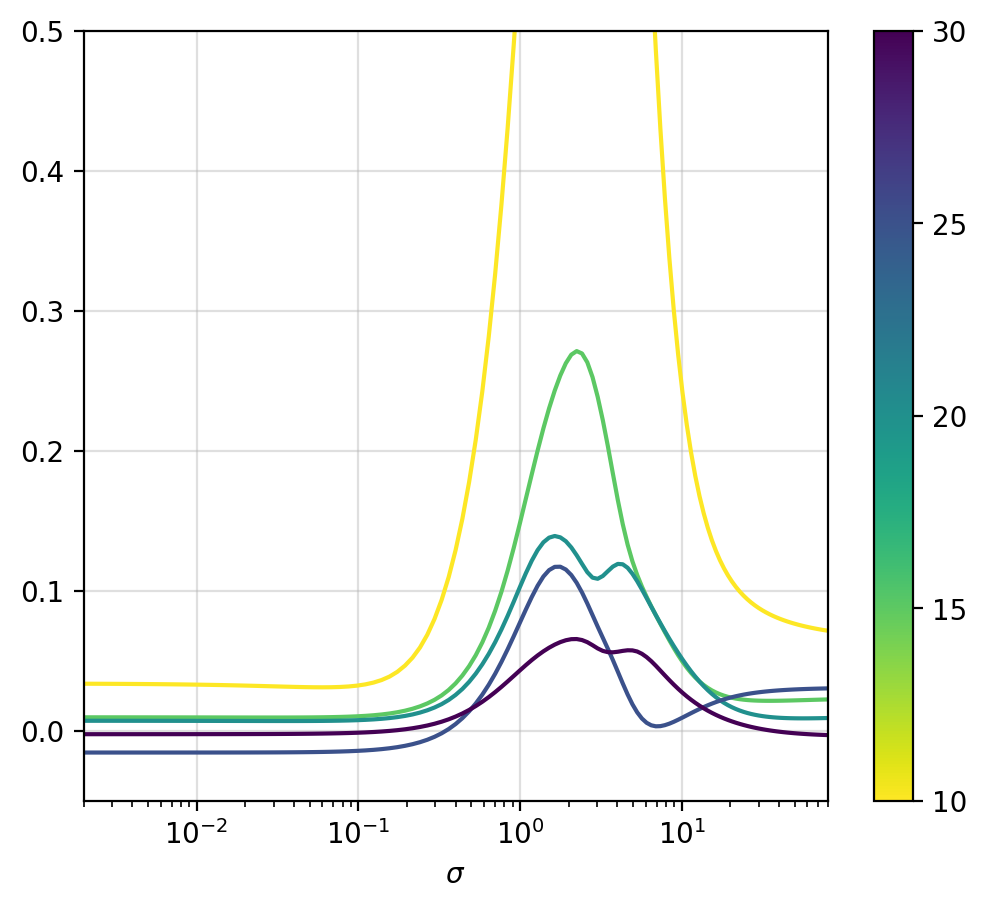}
         \caption{Training size $N$}
     \end{subfigure}
     \begin{subfigure}[b]{0.32\textwidth}
         \centering
         \includegraphics[scale=0.35]{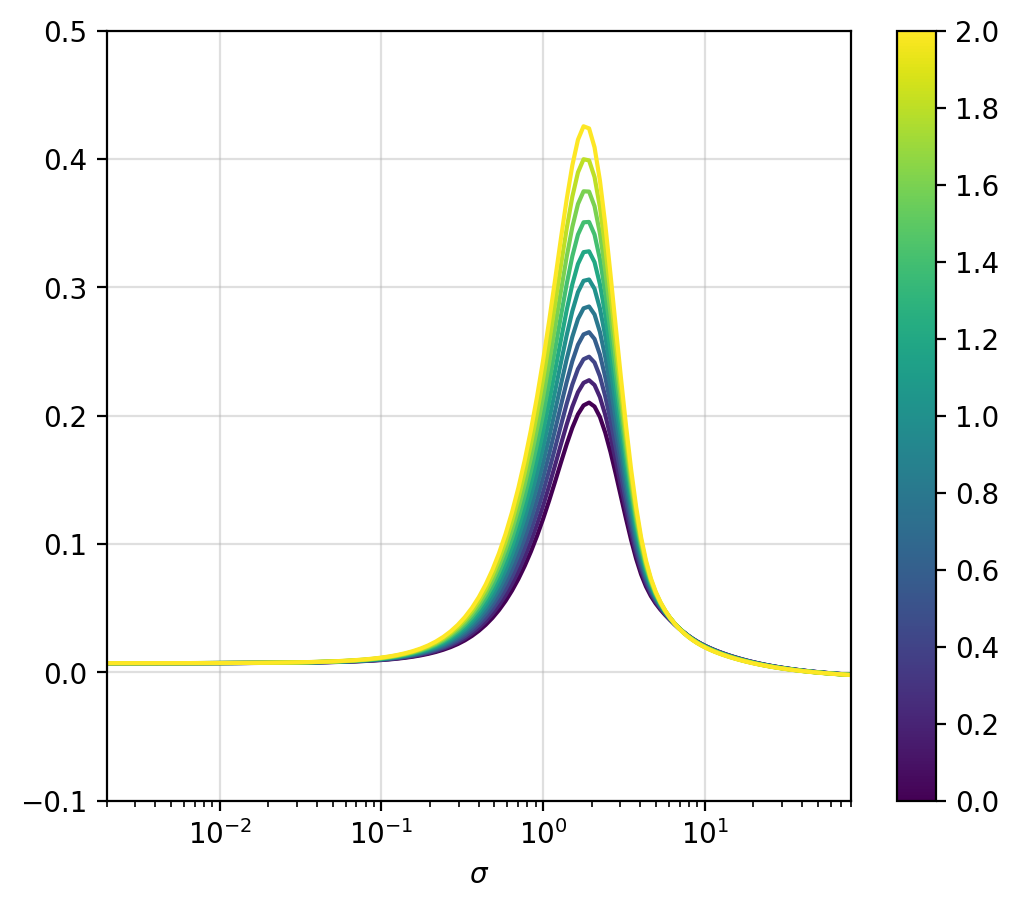}
         \caption{Guidance weight $w$}
     \end{subfigure}
     \caption{Relative generalization gap (\cref{eq:gen-gap}) with $M_\text{L2}$ in our 2D toy model with random data on a circular manifold and \textbf{(a)} different settings of the model error parameter $\delta$, \textbf{(b)} different number of training points, and \textbf{(c)} different settings for the guidance weight $w$. $w=0$ corresponds to no guidance.}
     \label{fig:toy-results-rnd}
\end{figure}

Instead of considering a symmetric distribution of $N$ training points $\{\y_i^{train}\}_{i=1}^N$ and $N$ validation points $\{\y_i^{val}\}_{i=1}^N$, we consider a random distribution of those points on a circle. The results are qualitatively the same, although considerably noisier, depending on the random seed. \cref{fig:toy-geometry-rnd} shows the flow field geometry for this setup. Note that, especially for small $N$, the superposition of all training points does not necessarily lie at the center of the circle (\cref{fig:toy-geometry-rnd}a), possibly causing the right tail of the generalization gap to be different from 0 (\cref{fig:toy-results-rnd})b. The manifold approximated by reconstructions at intermediate noise can have disconnected regions (\cref{fig:toy-geometry-rnd}b), where support is sparse, but densely populated regions collapse onto the training points later (\cref{fig:toy-geometry-rnd}c). The supplementary material includes a Jupyter Notebook that reproduces all figures related to the toy model and more.

\section{Additional results for the relative generalization gap} \label{supp:additional-gengaps}
This section presents the relative generalization gaps across all our metrics with the EDM2 \citep{karras2024edmv2} model on ImageNet-64/512 and the EDM \citep{karras2022edm} model on CIFAR-10/100. 

\clearpage
\subsection{ImageNet-64}
\vfill
\begin{figure}[h]
    \centering
    % Row 1: xs-uncond, xs, s
    \begin{subfigure}[b]{0.32\textwidth}
        \raggedright
        \includegraphics[width=\textwidth]{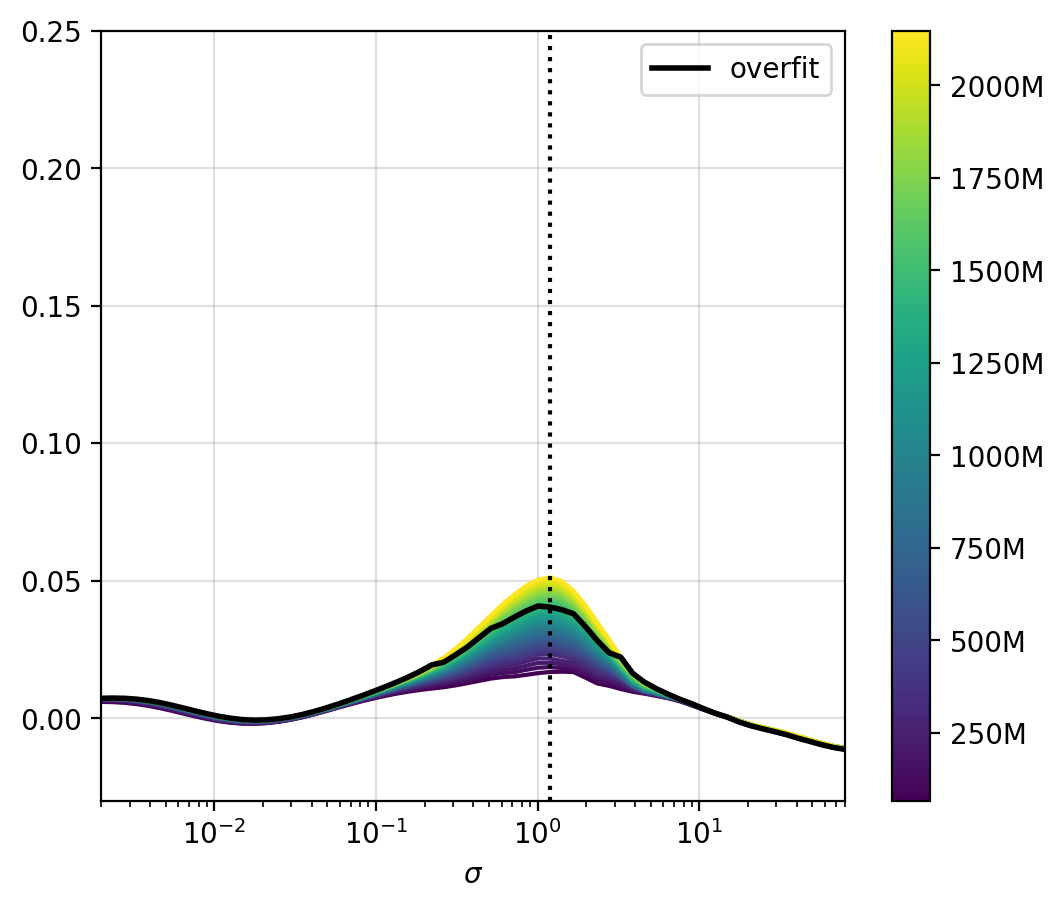}
        \caption{xs-uncond}
    \end{subfigure}
    \begin{subfigure}[b]{0.32\textwidth}
        \raggedright
        \includegraphics[width=\textwidth]{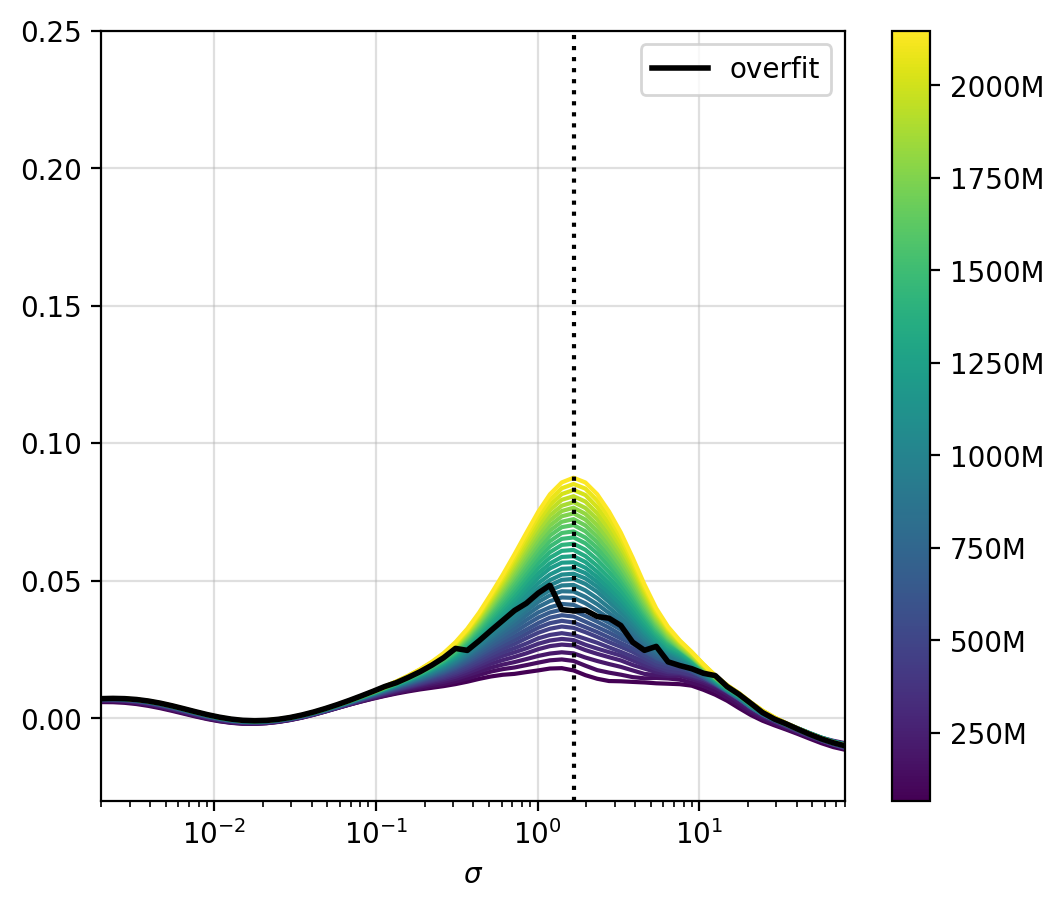}
        \caption{xs}
    \end{subfigure}
    \begin{subfigure}[b]{0.32\textwidth}
        \raggedright
        \includegraphics[width=\textwidth]{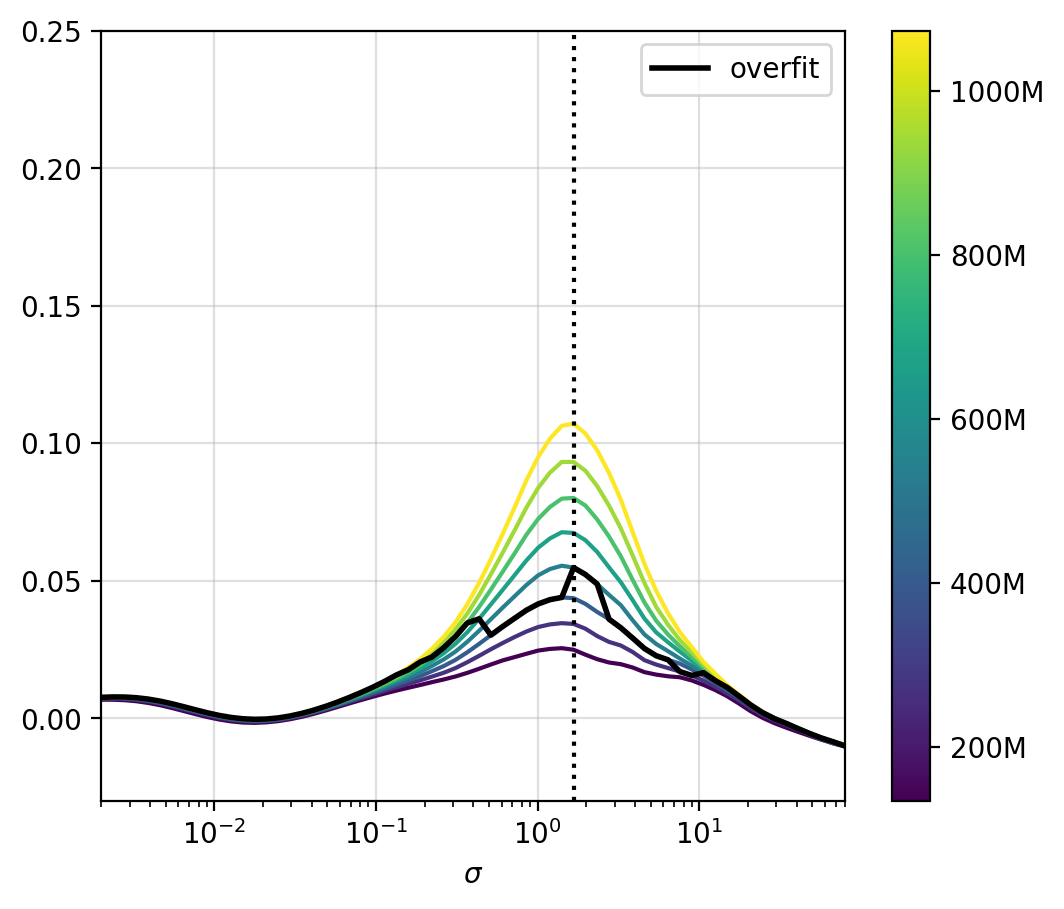}
        \caption{s}
    \end{subfigure}

    % Row 2: m, l, xl
    \begin{subfigure}[b]{0.32\textwidth}
        \raggedright
        \includegraphics[width=\textwidth]{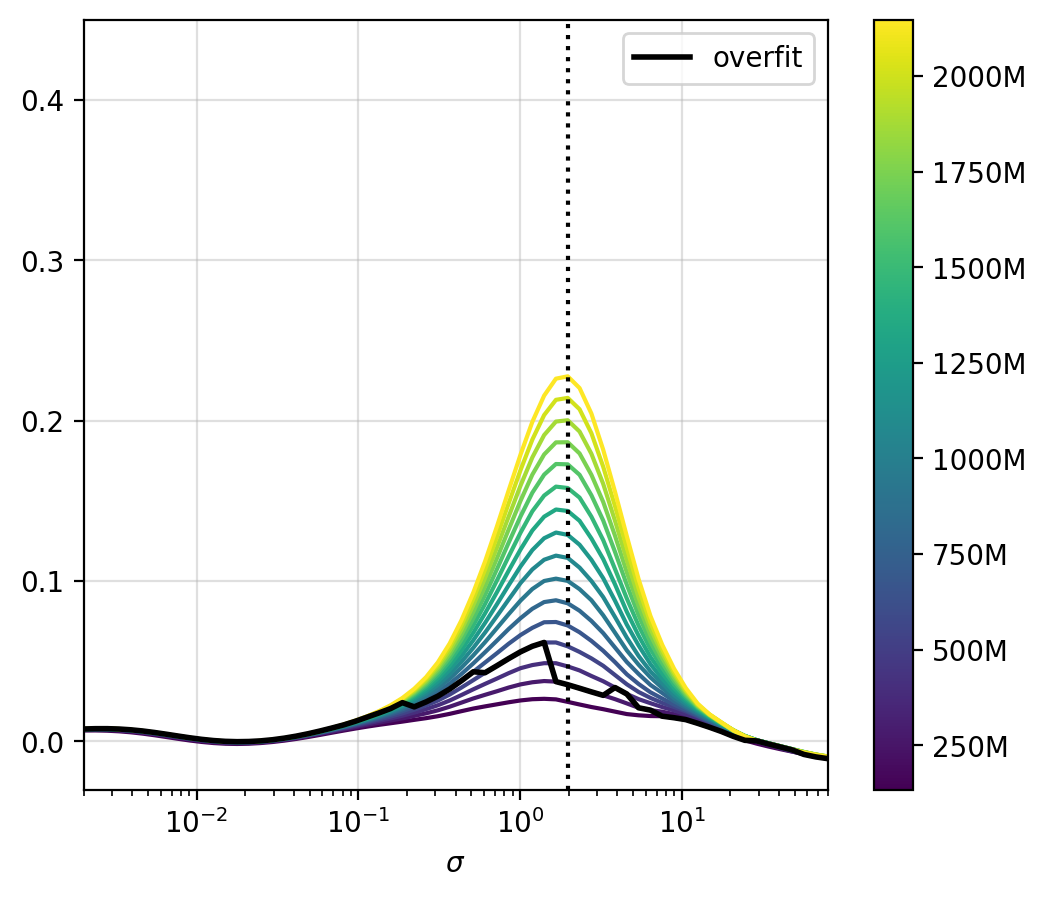}
        \caption{m}
    \end{subfigure}
    \begin{subfigure}[b]{0.32\textwidth}
        \raggedright
        \includegraphics[width=\textwidth]{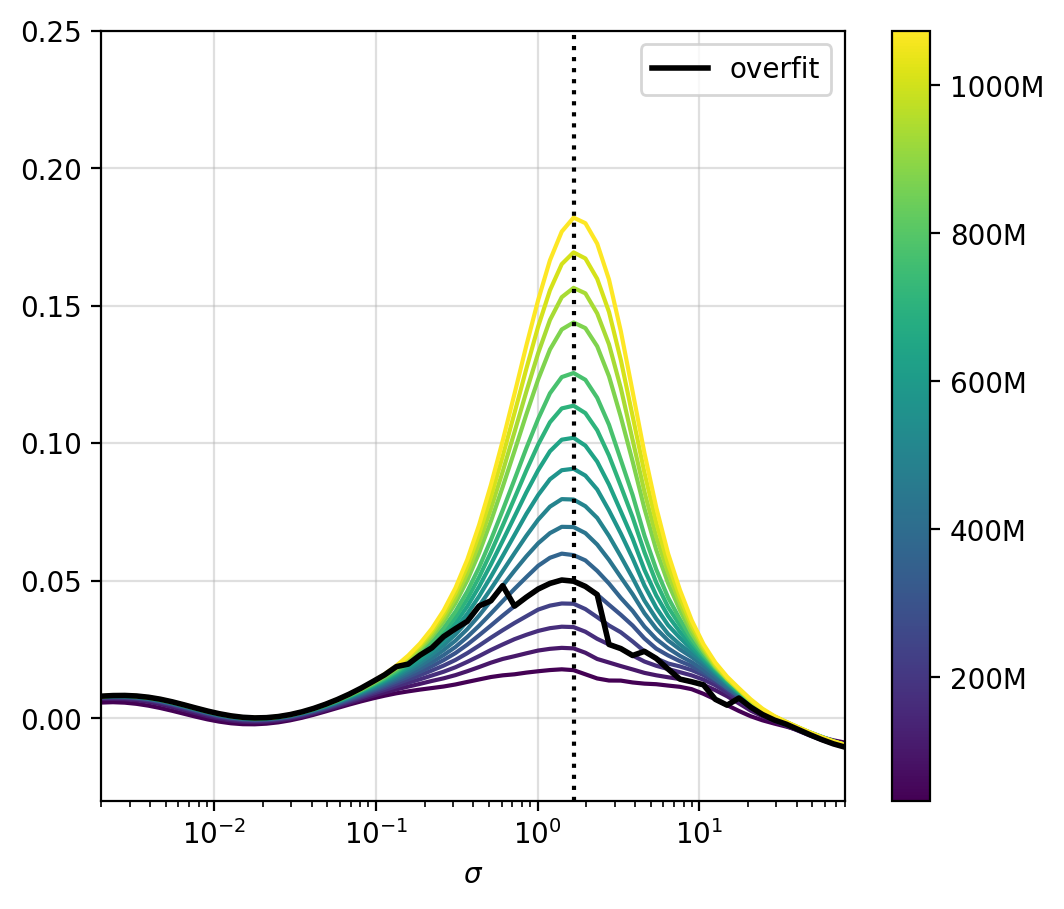}
        \caption{l}
    \end{subfigure}
    \begin{subfigure}[b]{0.32\textwidth}
        \raggedright
        \includegraphics[width=\textwidth]{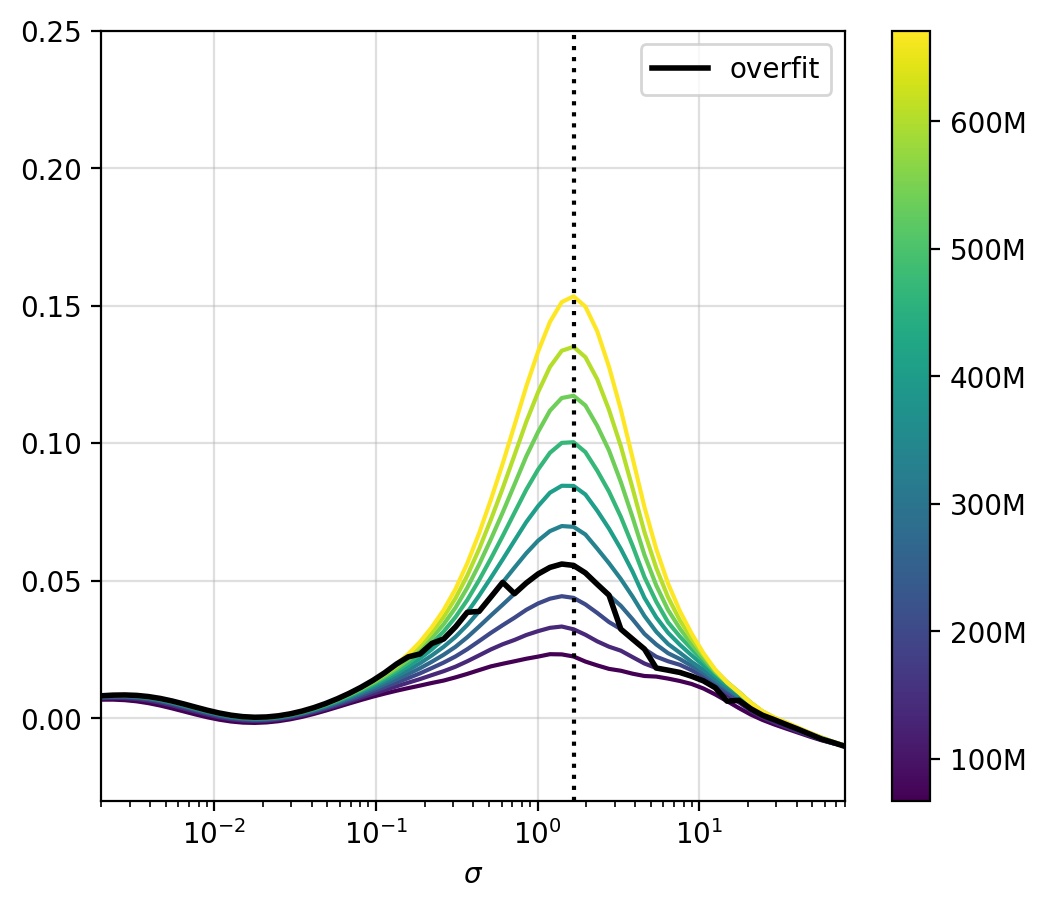}
        \caption{xl}
    \end{subfigure}
    \caption{Relative generalization gap (\cref{eq:gen-gap}) with EDM2 on ImageNet-64 for various model sizes, plotted vs $\sigma$. Colorbar shows images seen during training. Dotted black lines indicate $\sigma$ values used in \cref{fig:in64-pl2-model-sizes-comp}.}
    \label{fig:in64-pl2-sigma-model-sizes}
\end{figure}
\vfill
\begin{figure}[h]
    \centering
    % Row 1: xs-uncond, xs, s
    \begin{subfigure}[b]{0.45\textwidth}
        \centering
        \includegraphics[width=\textwidth]{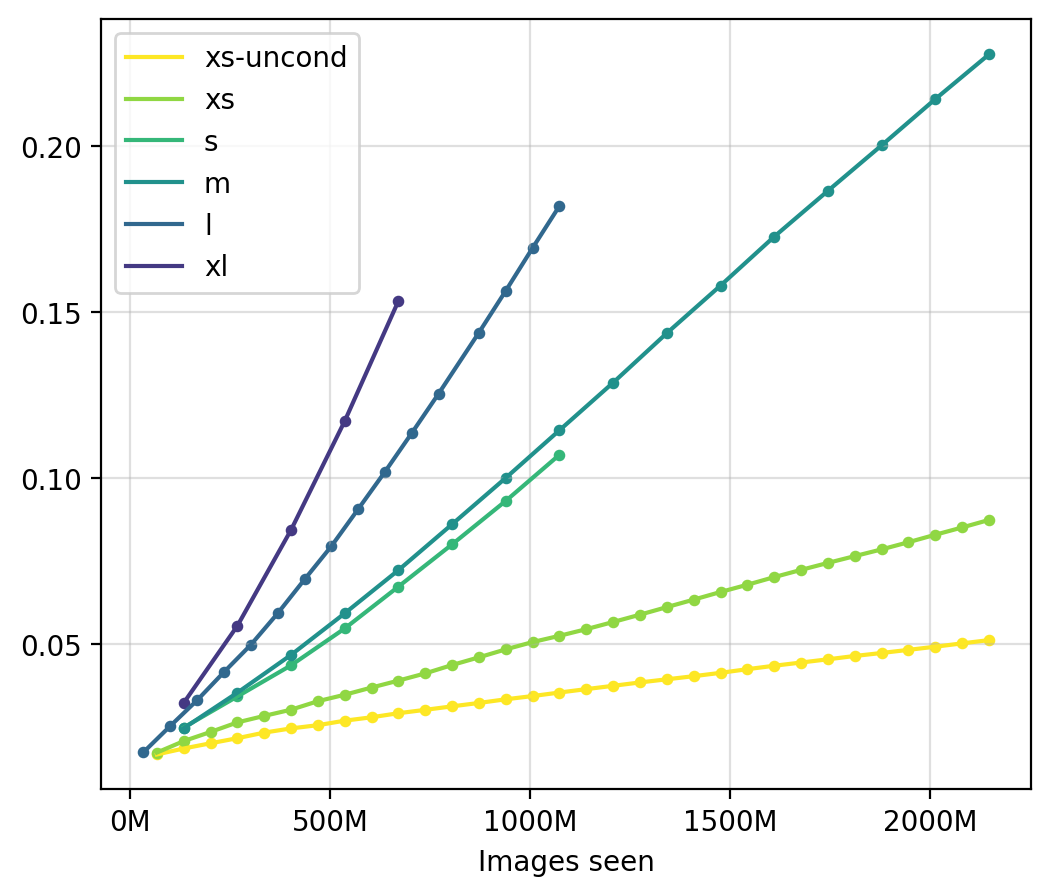}
        \caption{xs-uncond}
    \end{subfigure}
    \hspace{5mm}
    \begin{subfigure}[b]{0.45\textwidth}
        \centering
        \includegraphics[width=\textwidth]{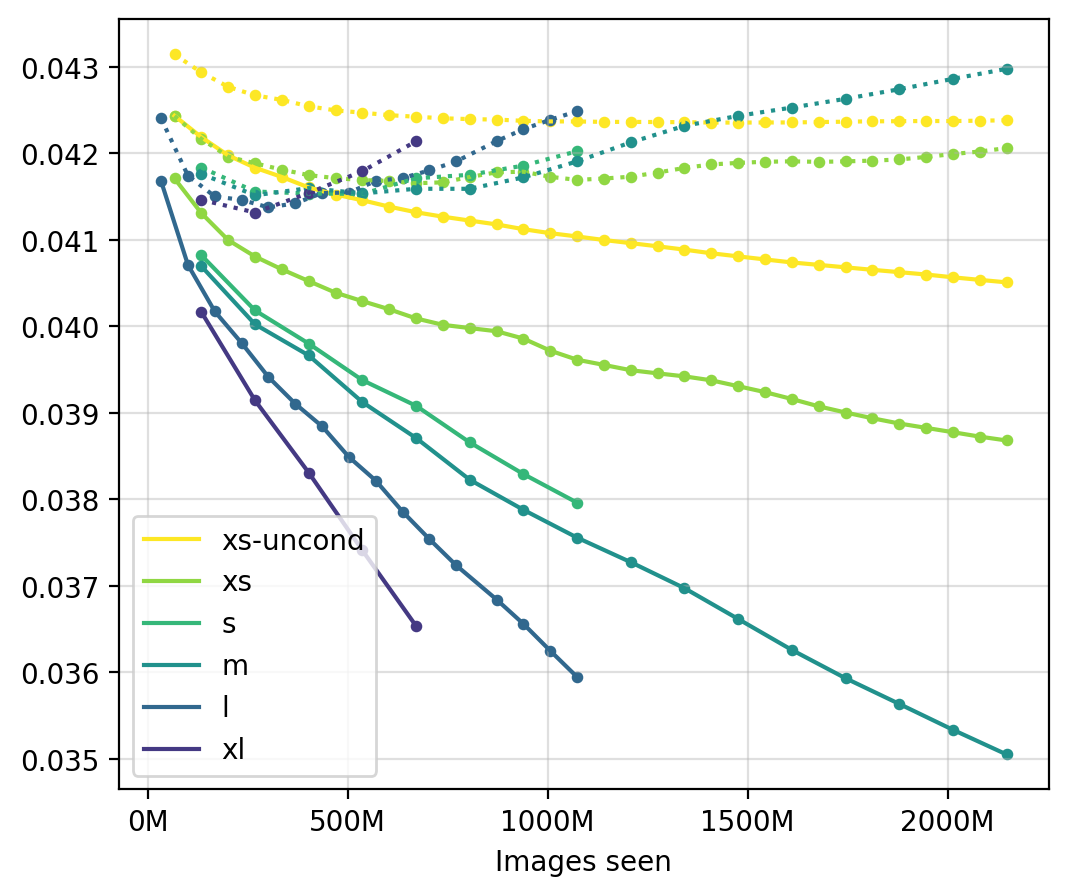}
        \caption{xs}
    \end{subfigure}
    \caption{Relative generalization gap (\cref{eq:gen-gap}) and training and validation results, with EDM2 on ImageNet-64 for various model sizes. For each model size, $\sigma$ is fixed at the peak of the generalization gap in \cref{fig:in64-pl2-sigma-model-sizes}.}
    \label{fig:in64-pl2-model-sizes-comp}
\end{figure}
\vfill
\clearpage

\subsection{ImageNet-512}
\vfill
\begin{figure}[h]
    \raggedright
    \begin{subfigure}[b]{0.32\textwidth}
        \raggedright
        \includegraphics[width=\textwidth]{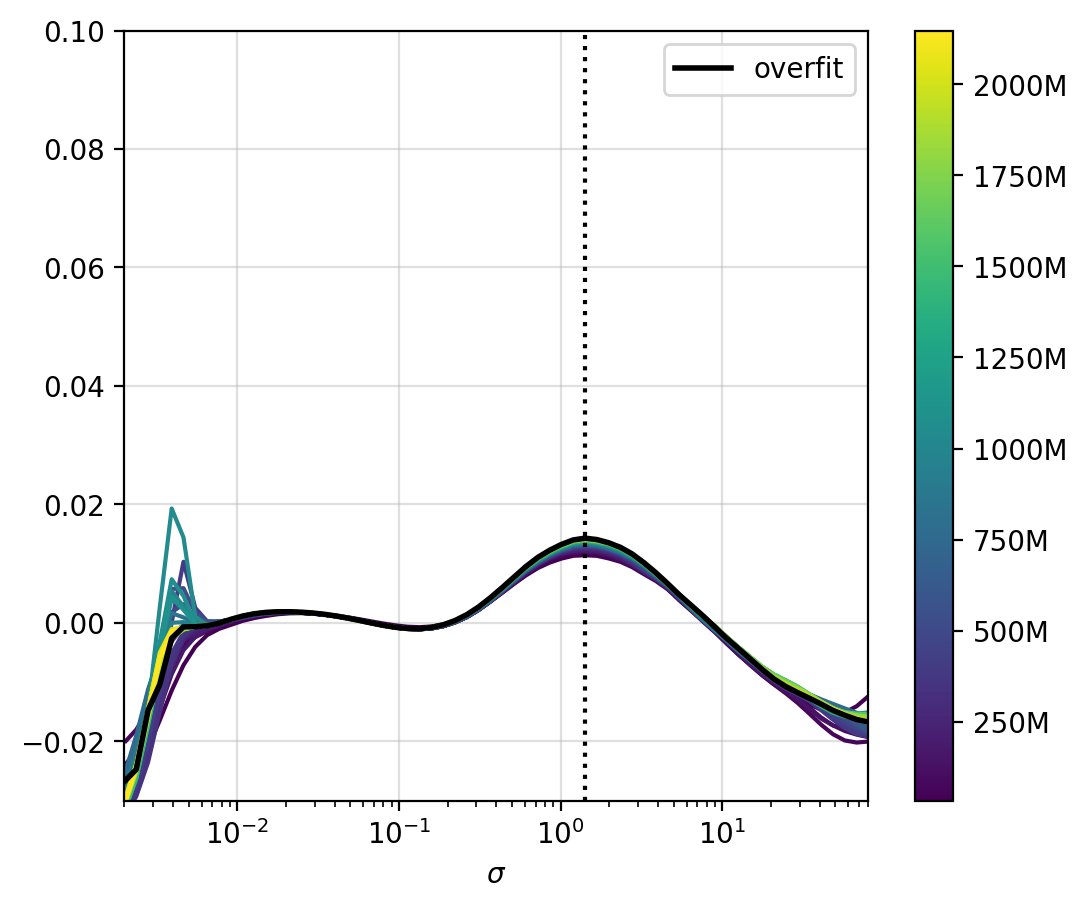}
        \caption{xss}
    \end{subfigure}
    \begin{subfigure}[b]{0.32\textwidth}
        \raggedright
        \includegraphics[width=\textwidth]{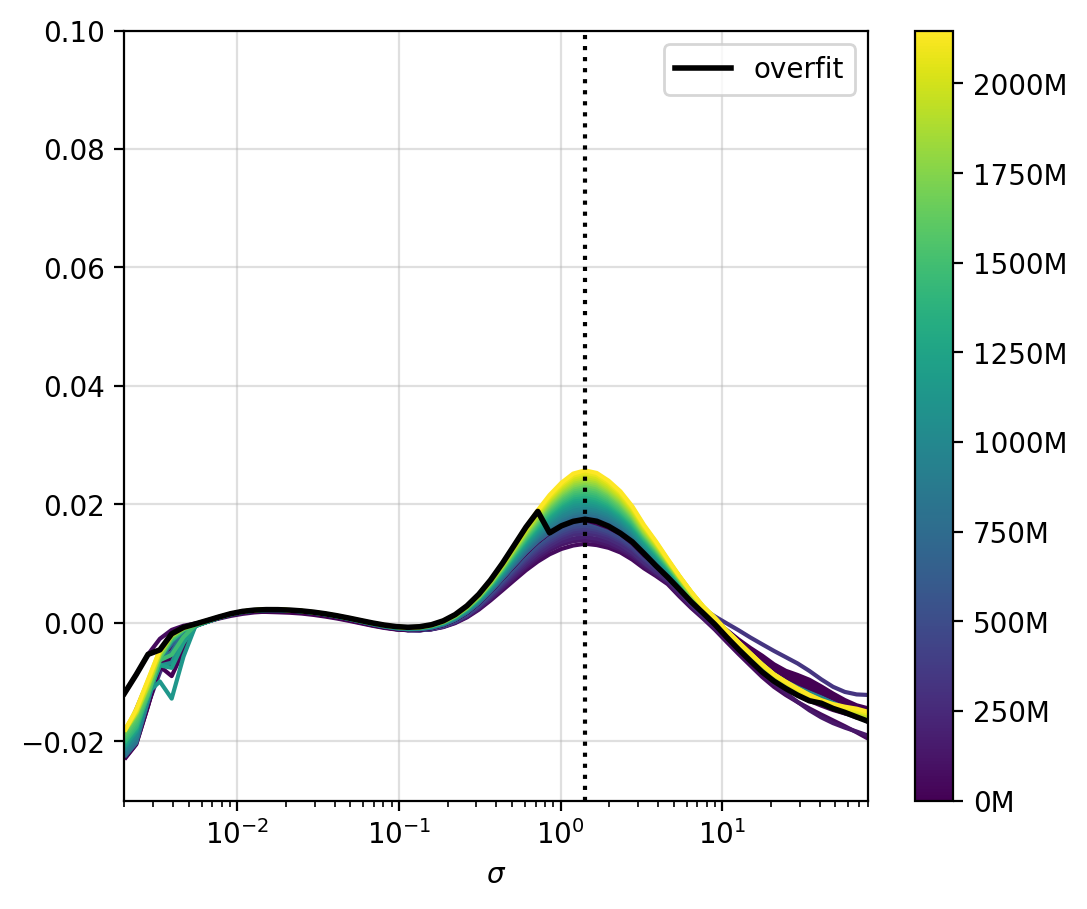}
        \caption{xs}
    \end{subfigure}
     \begin{subfigure}[b]{0.32\textwidth}
        \raggedright
        \includegraphics[width=\textwidth]{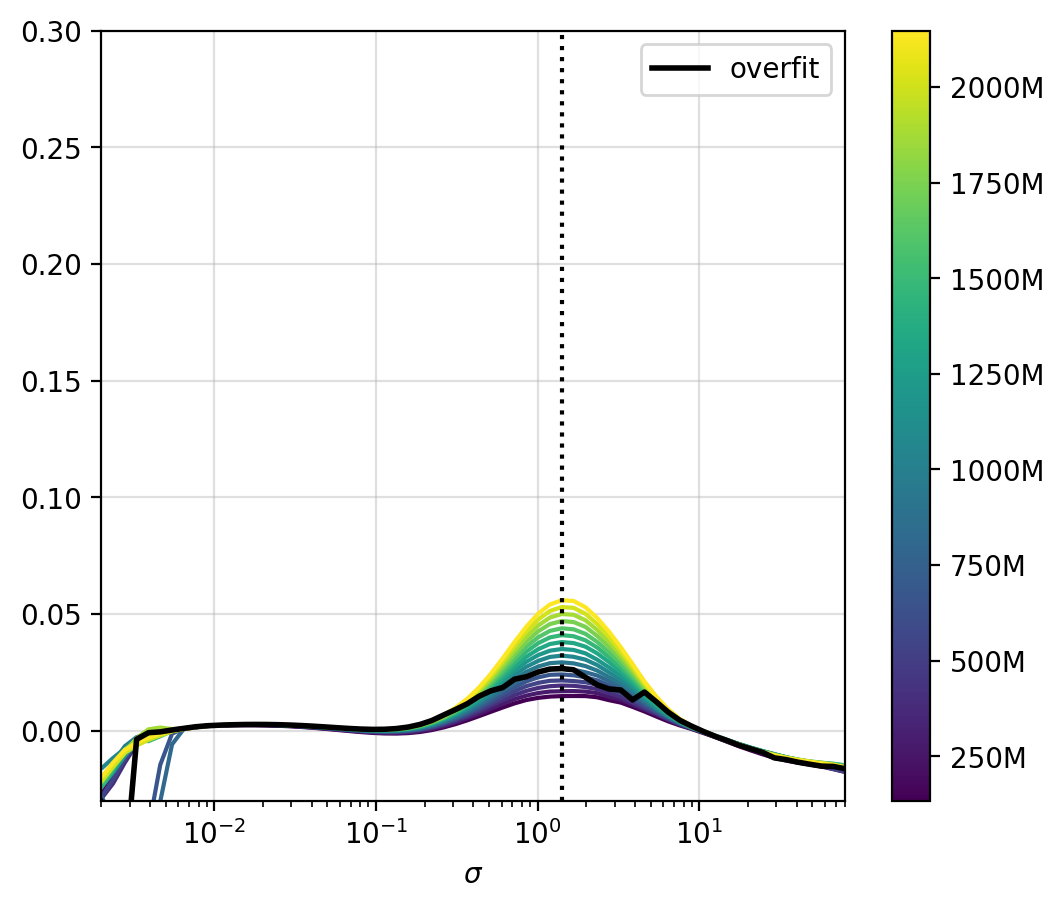}
        \caption{s}
    \end{subfigure}
    
    \vspace{3mm}
     
    \begin{subfigure}[b]{0.32\textwidth}
        \raggedright
        \includegraphics[width=\textwidth]{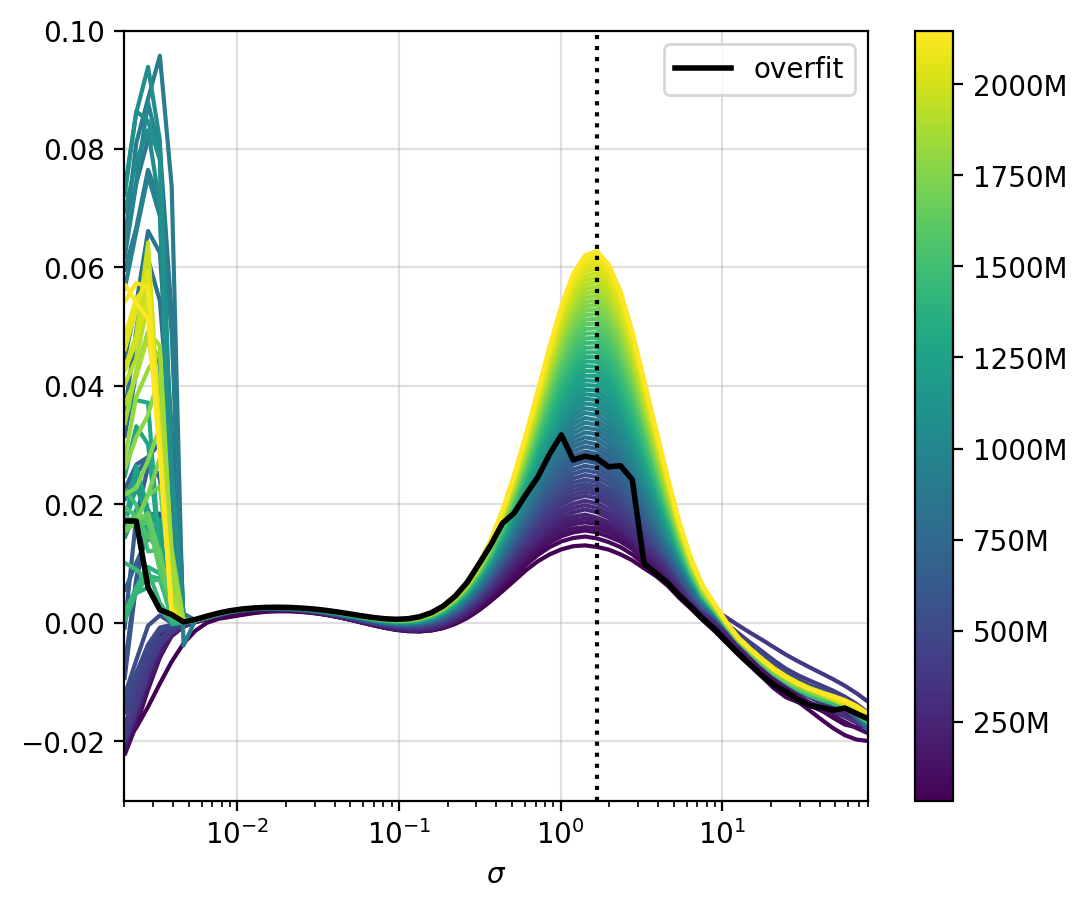}
        \caption{m}
    \end{subfigure}
    \begin{subfigure}[b]{0.32\textwidth}
        \raggedright
        \includegraphics[width=\textwidth]{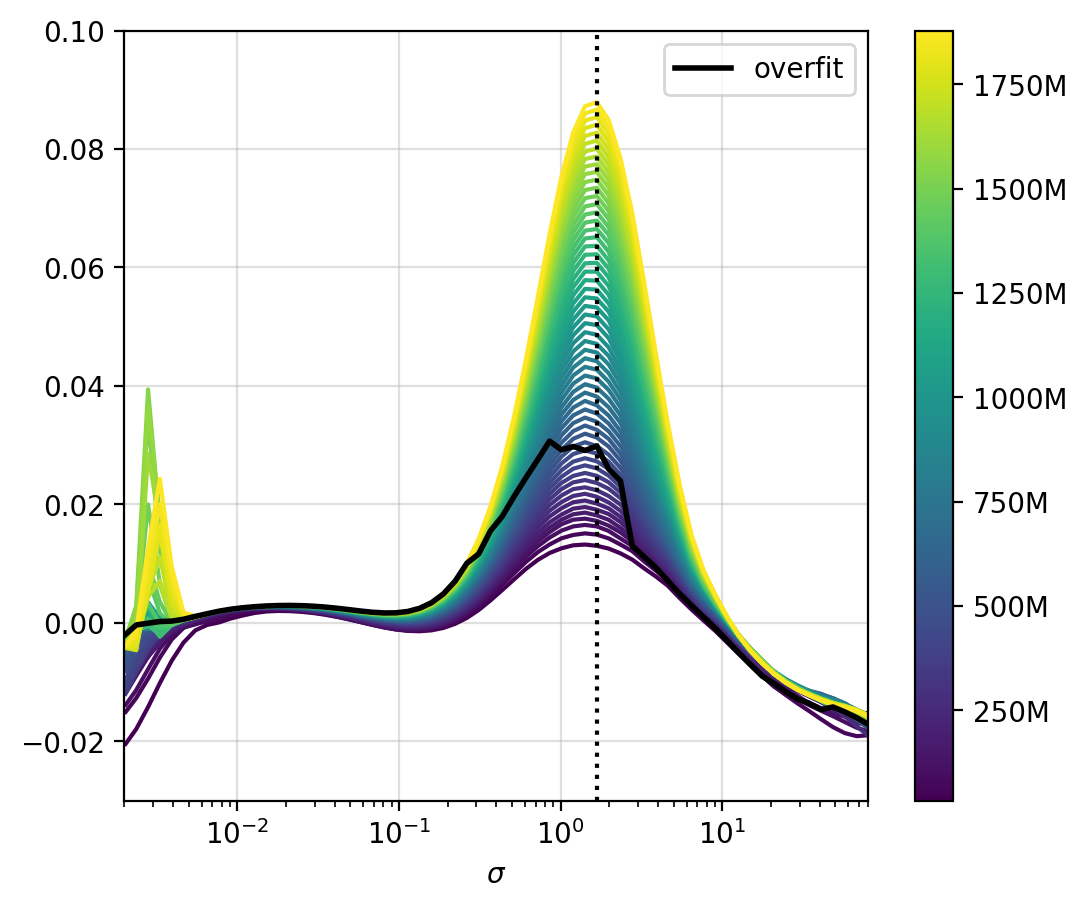}
        \caption{l}
    \end{subfigure}
    \begin{subfigure}[b]{0.32\textwidth}
        \raggedright
        \includegraphics[width=\textwidth]{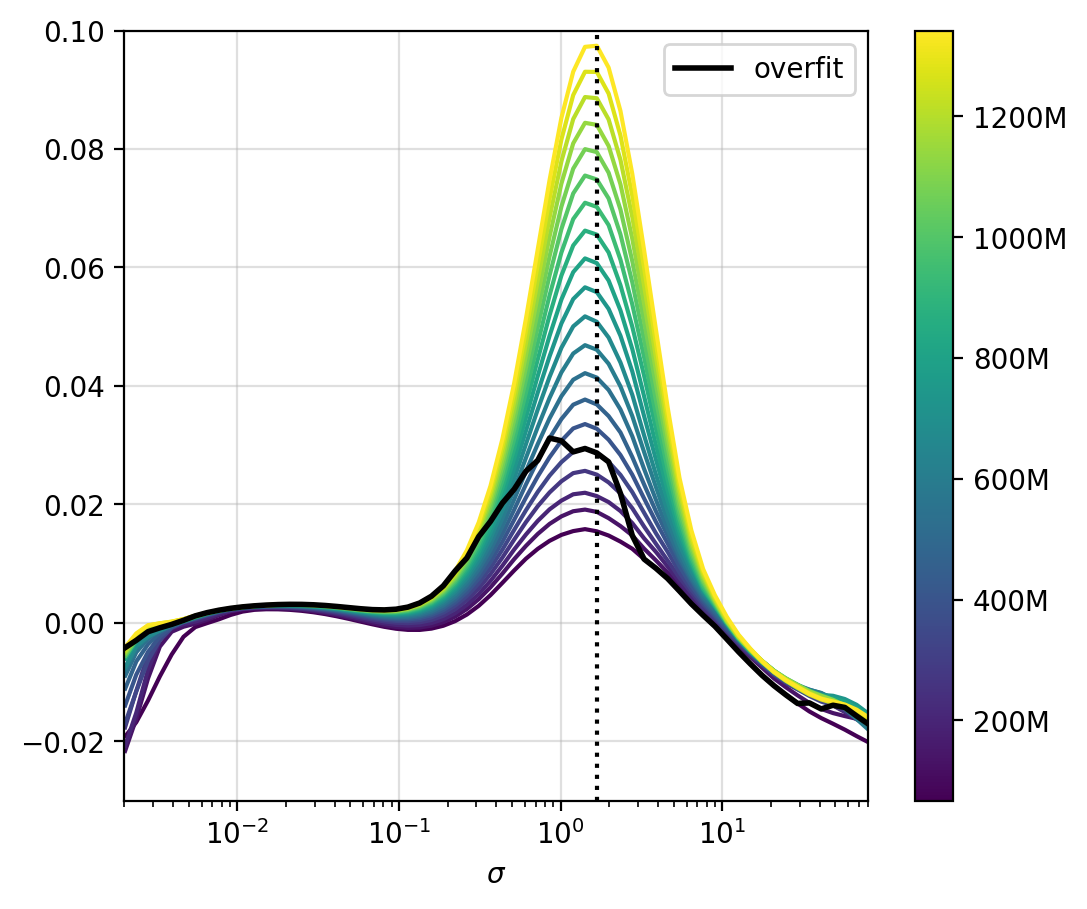}
        \caption{xl}
    \end{subfigure}
    
    \caption{Relative generalization gap (\cref{eq:gen-gap}) with EDM2 on ImageNet-512 for various model sizes, plotted against the noise level $\sigma$. Colorbar shows images seen during training. Dotted black lines indicate $\sigma$ values used in \cref{fig:in512-pl2-model-sizes-comp}.}
    \label{fig:in512-pl2-sigma-model-sizes}
\end{figure}
\vspace*{\fill}
\vfill
\begin{figure}[h]
    \centering
    % Row 1: xs-uncond, xs, s
    \begin{subfigure}[b]{0.45\textwidth}
        \centering
        \includegraphics[width=\textwidth]{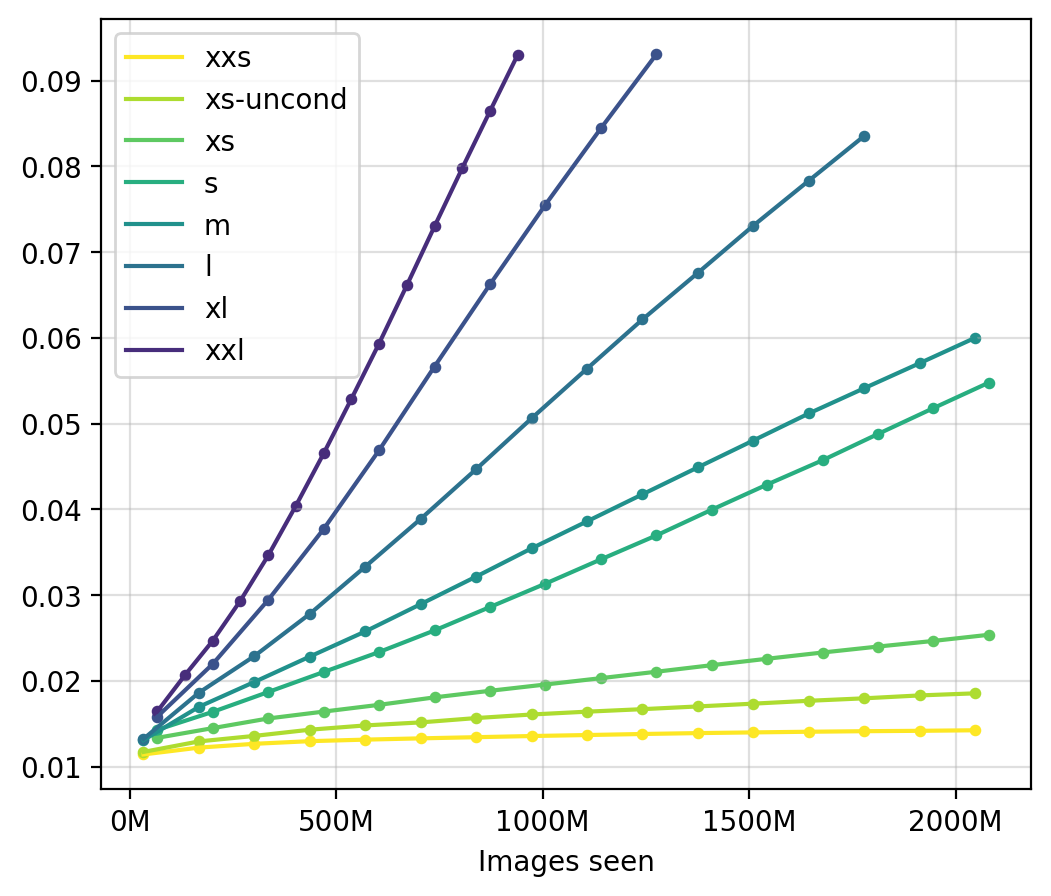}
        \caption{xs-uncond}
    \end{subfigure}
    \hspace{5mm}
    \begin{subfigure}[b]{0.45\textwidth}
        \centering
        \includegraphics[width=\textwidth]{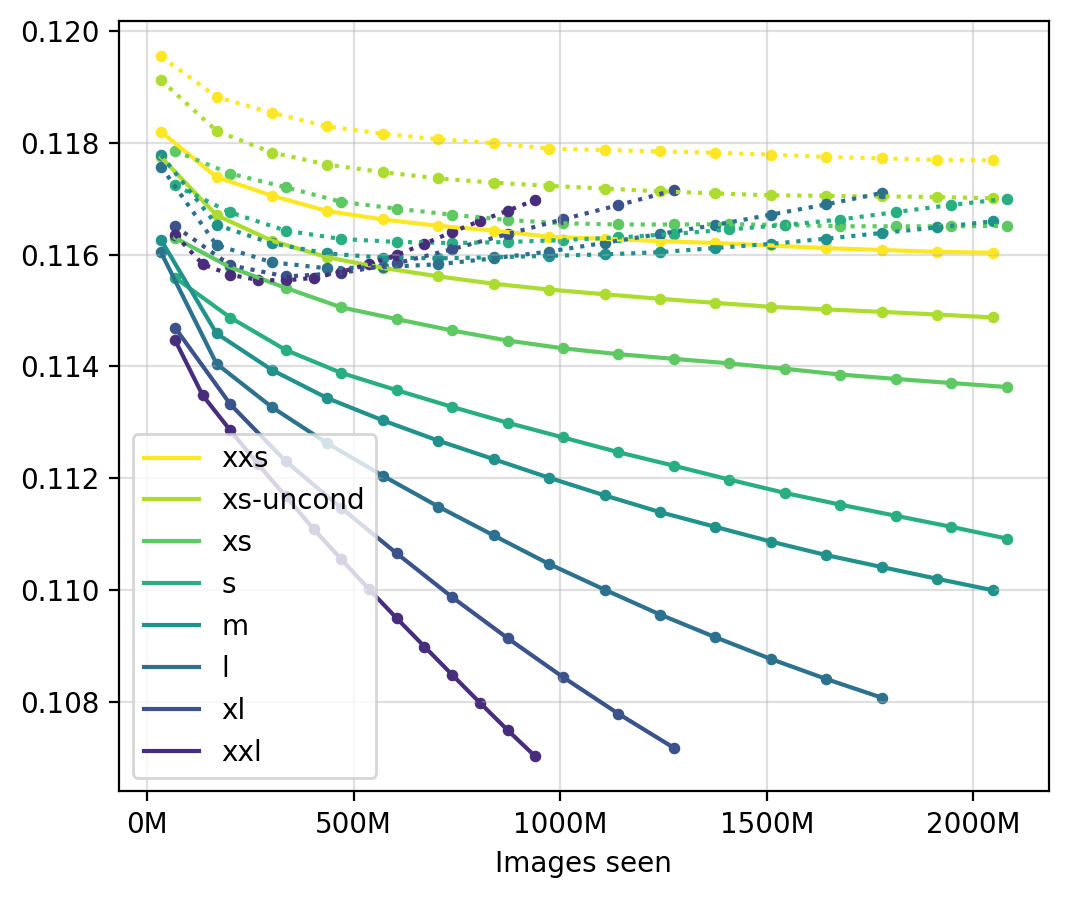}
        \caption{xs}
    \end{subfigure}
    \caption{Relative generalization gap (\cref{eq:gen-gap}) and training and validation results, with EDM2 on ImageNet-64 for various model sizes. For each model size, $\sigma$ is fixed at the peak of the generalization gap in \cref{fig:in512-pl2-sigma-model-sizes}.}
    \label{fig:in512-pl2-model-sizes-comp}
\end{figure}
\vfill
\clearpage

\subsection{CIFAR-10/100}
\vfill
\begin{figure}[h]
    \centering
    \begin{subfigure}[b]{0.45\textwidth}
        \centering
        \includegraphics[width=\textwidth]{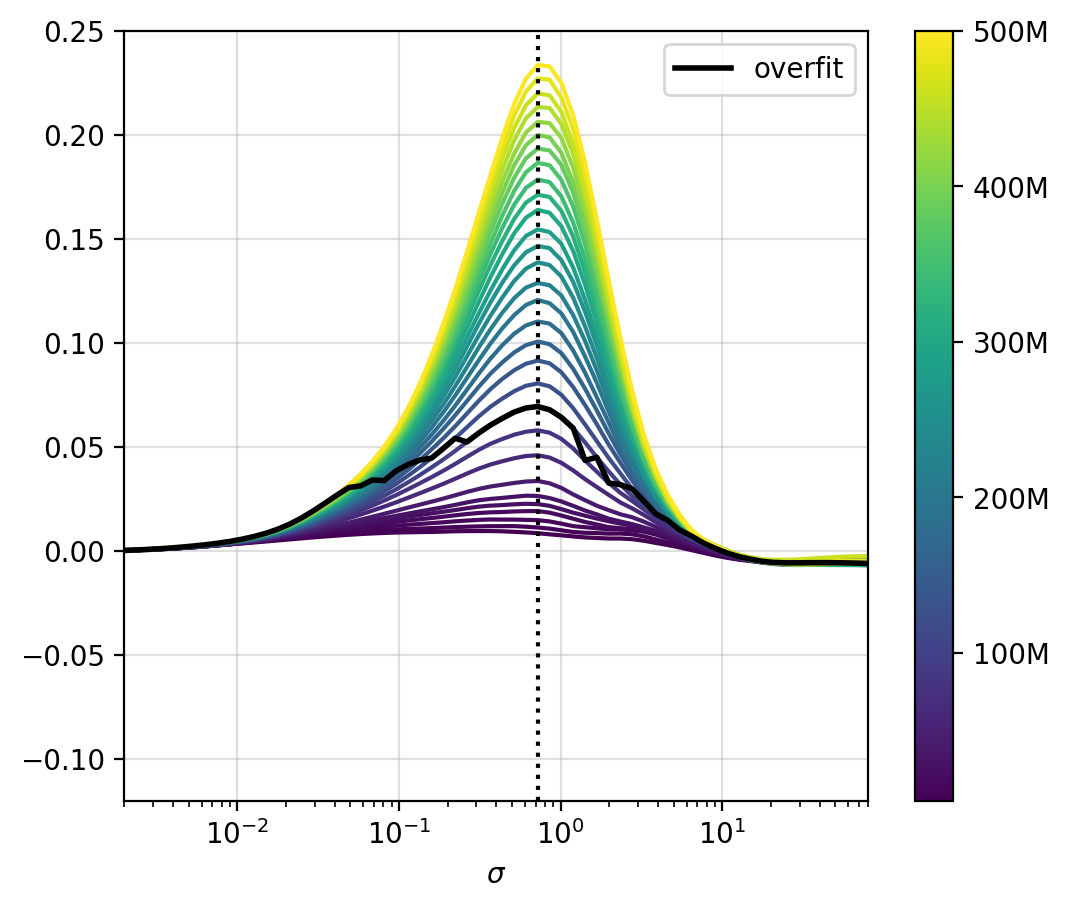}
        \caption{CIFAR-10}
    \end{subfigure}
    \hspace{5mm}
    \begin{subfigure}[b]{0.45\textwidth}
        \centering
        \includegraphics[width=\textwidth]{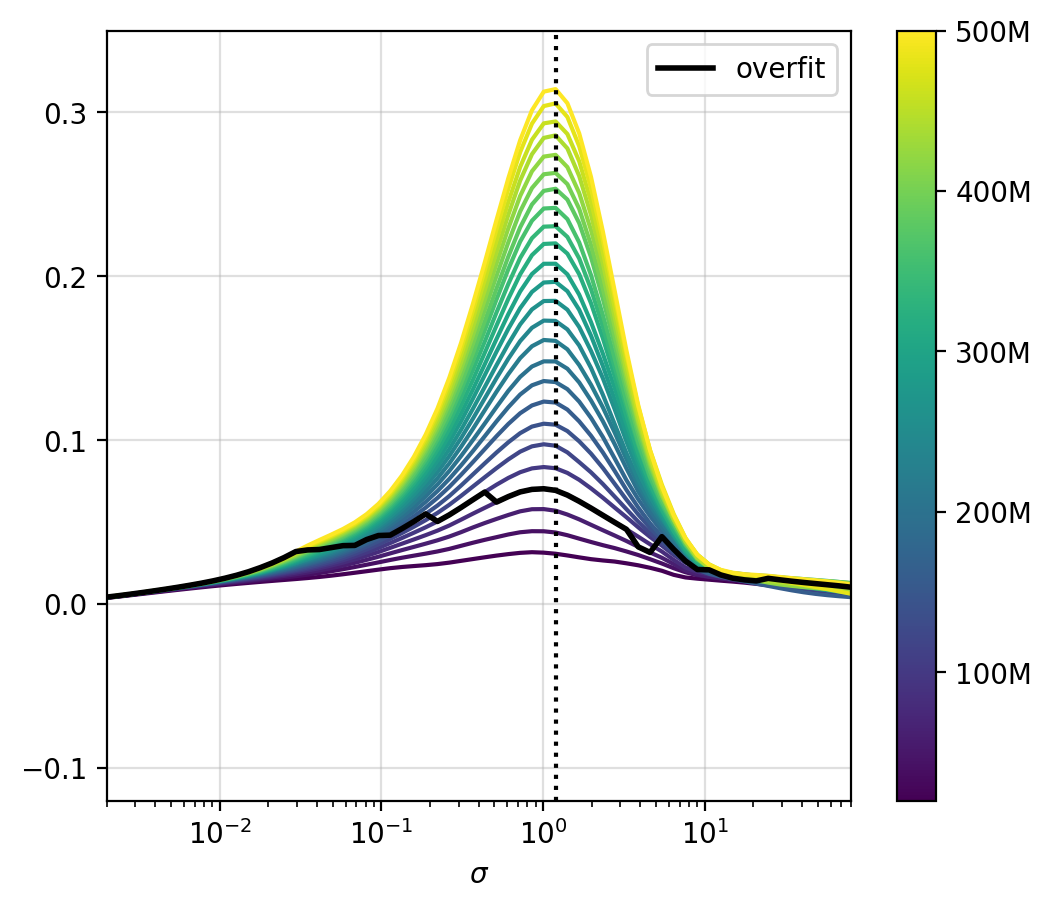}
        \caption{CIFAR-100}
    \end{subfigure}
    \caption{Relative generalization gap (\cref{eq:gen-gap}) with EDM on CIFAR-10/100. Colorbar shows images seen during training. Dotted black lines indicate $\sigma$ values used in \cref{fig:cifar-metrics-model-error-gap}}
    \label{fig:cifar-edm-metrics-sigma}
\end{figure}
\vfill
\begin{figure}[h]
    \centering
    \begin{subfigure}[b]{0.45\textwidth}
        \centering
        \includegraphics[width=\textwidth]{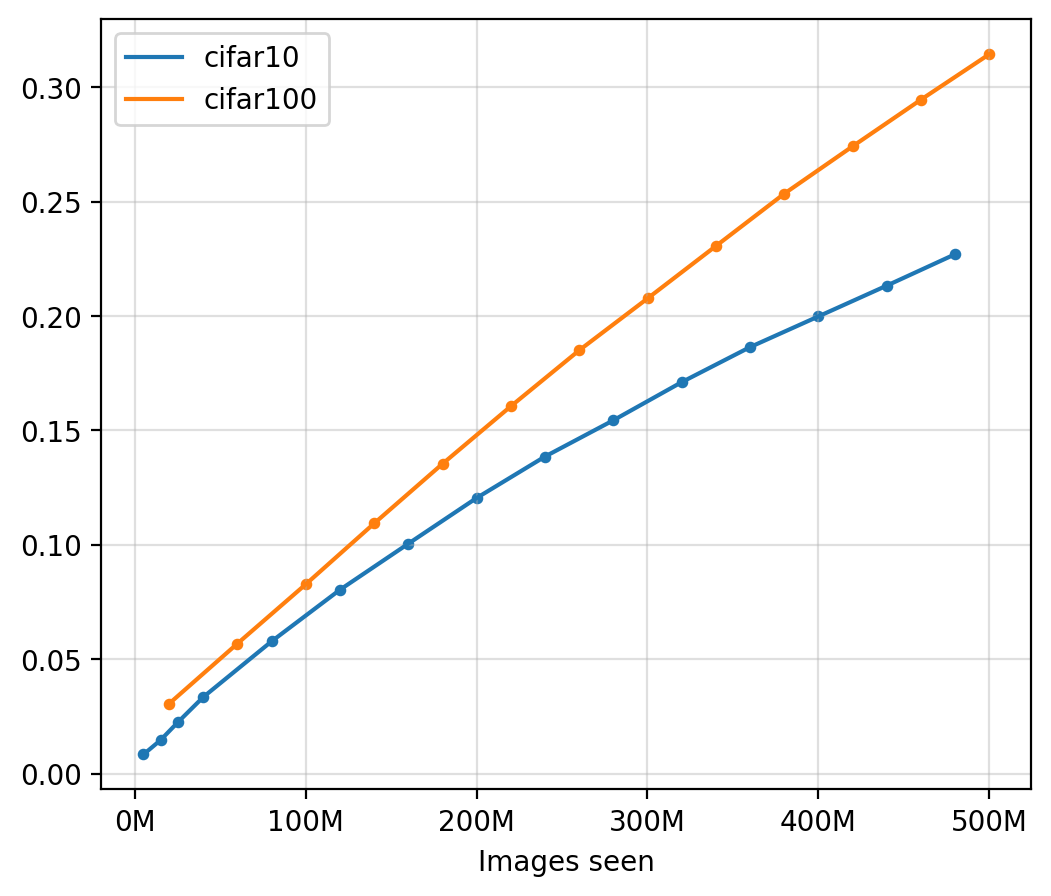}
        \caption{Rel. gen-gap}
    \end{subfigure}
    \hspace{5mm}
    \begin{subfigure}[b]{0.45\textwidth}
        \centering
        \includegraphics[width=\textwidth]{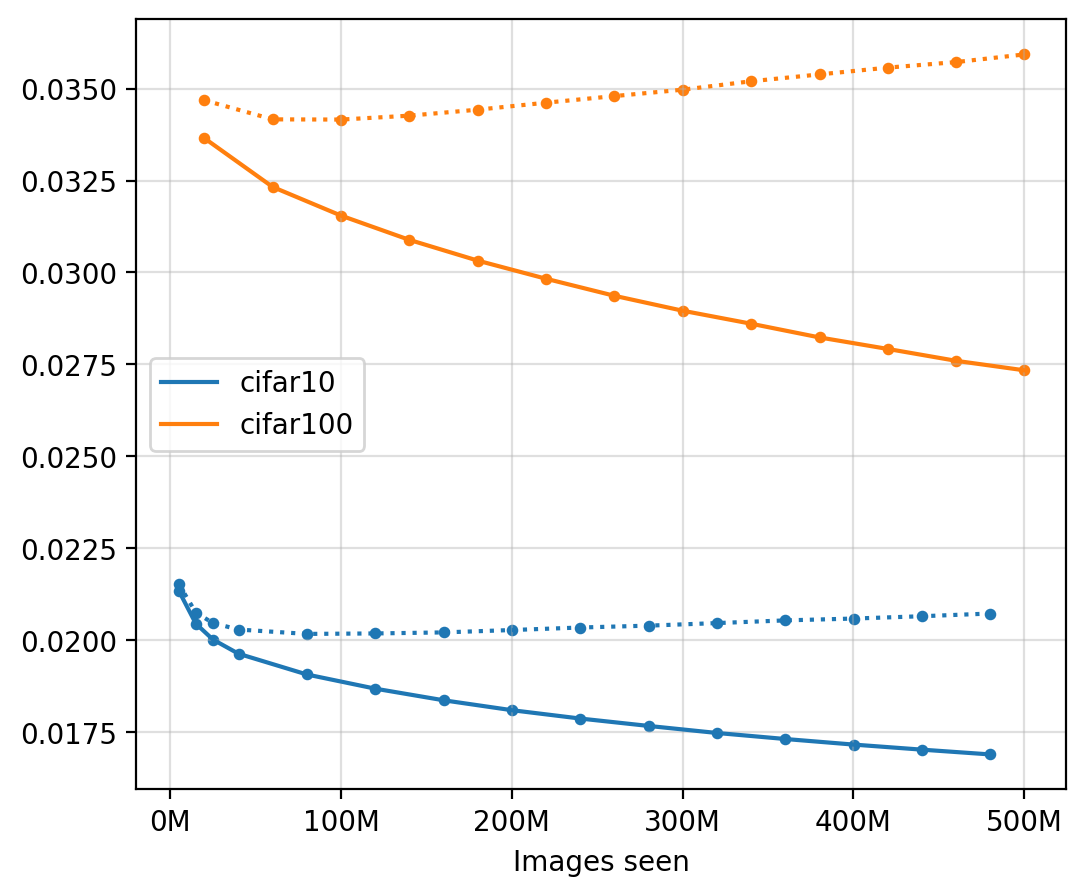}
        \caption{Train/val results}
    \end{subfigure}
    \caption{Relative generalization gap (\cref{eq:gen-gap}) and training and validation results with EDM on CIFAR-10/100 for model sizes. $\sigma$ is fixed at the peak of the generalization gap, see \cref{fig:cifar-edm-metrics-sigma}.}
    \label{fig:cifar-metrics-model-error-gap}
\end{figure}
\vfill

\clearpage
\section{Additional results for the similarity metrics} \label{supp:additional-knn}

\vspace*{\fill}
\begin{figure}[h]
    \centering
    \begin{subfigure}[t]{0.32\textwidth}
        \centering
        \includegraphics[scale=0.37]{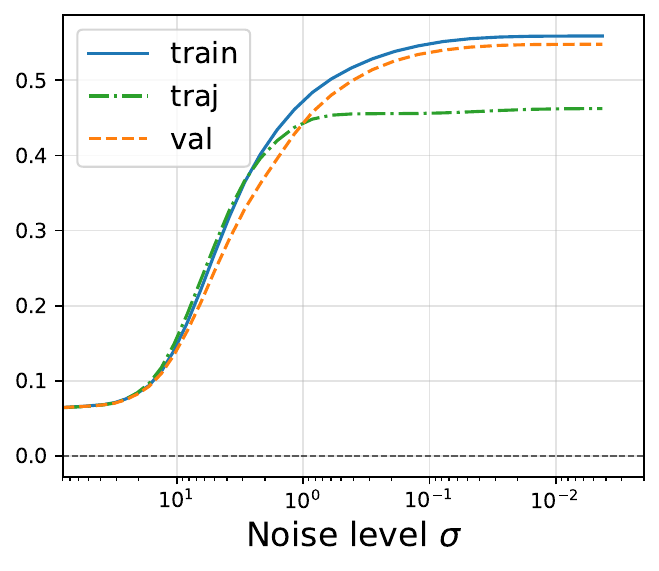}
        \caption{$S_{20}(Q, {\cal D}_\text{train};\sigma)$}
    \end{subfigure}
    \begin{subfigure}[t]{0.32\textwidth}
        \centering
        \includegraphics[scale=0.37]{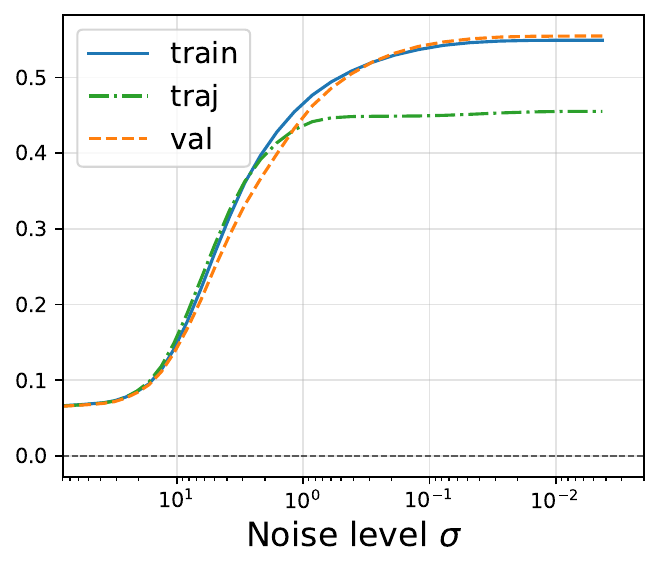}
        \caption{$S_{20}(Q, {\cal D}_\text{train};\sigma)$}
    \end{subfigure}
    \begin{subfigure}[t]{0.32\textwidth}
        \centering
        \includegraphics[scale=0.37]{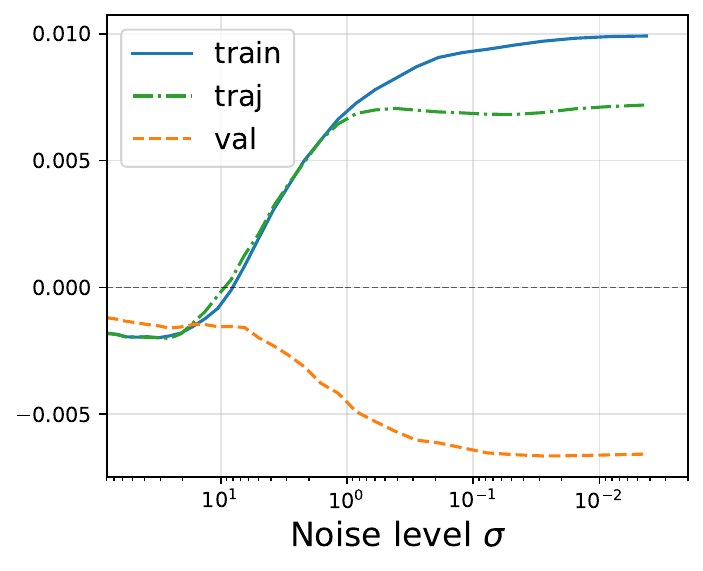}
        \caption{$B_{20}(Q; \sigma)$}
    \end{subfigure}
    
    % \vspace{3mm}
     
    % \begin{subfigure}[t]{0.32\textwidth}
    %     \centering
    %     \includegraphics[scale=0.37]{imgs/Q_vs_sigma_k20_in64_edm2-m_Sk_train_2147483.pdf}
    %     \caption{EDM2-M}
    % \end{subfigure}
    % \begin{subfigure}[t]{0.32\textwidth}
    %     \centering
    %     \includegraphics[scale=0.37]{imgs/Q_vs_sigma_k20_in64_edm2-m_Sk_val_2147483.pdf}
    %     \caption{EDM2-M}
    % \end{subfigure}
    % \begin{subfigure}[t]{0.32\textwidth}
    %     \centering
    %     \includegraphics[scale=0.37]{imgs/Q_vs_sigma_k20_in64_edm2-m_Bk_2147483.pdf}
    %     \caption{EDM2-M}
    % \end{subfigure}
    
    % \vspace{3mm}
     
    % \begin{subfigure}[t]{0.32\textwidth}
    %     \centering
    %     \includegraphics[scale=0.37]{imgs/Q_vs_sigma_k20_in64_edm2-l_Sk_train_0973078.pdf}
    %     \caption{EDM2-L}
    % \end{subfigure}
    % \begin{subfigure}[t]{0.32\textwidth}
    %     \centering
    %     \includegraphics[scale=0.37]{imgs/Q_vs_sigma_k20_in64_edm2-l_Sk_val_0973078.pdf}
    %     \caption{EDM2-L}
    % \end{subfigure}
    % \begin{subfigure}[t]{0.32\textwidth}
    %     \centering
    %     \includegraphics[scale=0.37]{imgs/Q_vs_sigma_k20_in64_edm2-l_Bk_0973078.pdf}
    %     \caption{EDM2-L}
    % \end{subfigure}

    \vspace{3mm}

    \begin{subfigure}[t]{0.32\textwidth}
        \centering
        \includegraphics[scale=0.37]{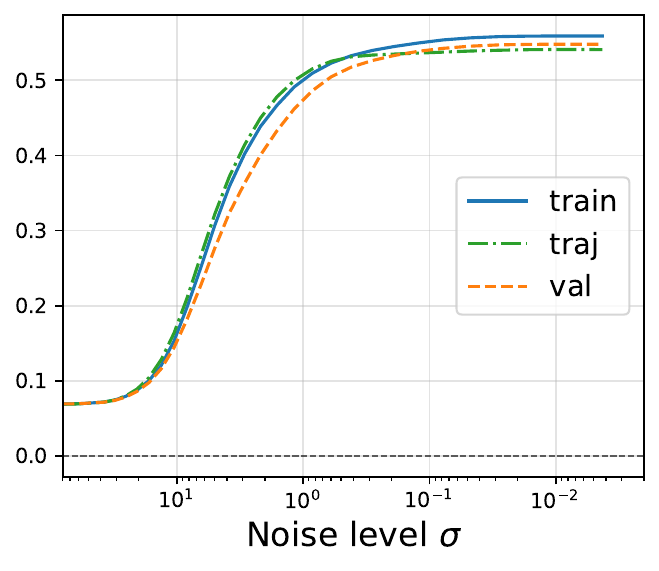}
        \caption{$S_{20}(Q, {\cal D}_\text{train};\sigma)$}
    \end{subfigure}
    \begin{subfigure}[t]{0.32\textwidth}
        \centering
        \includegraphics[scale=0.37]{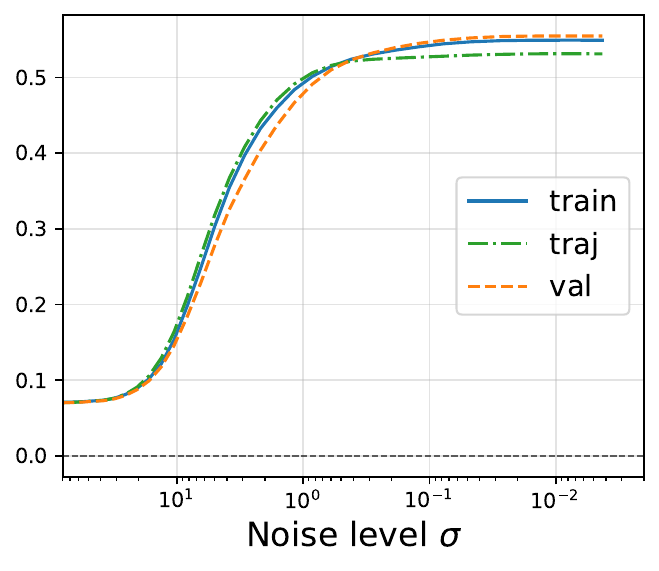}
        \caption{$S_{20}(Q, {\cal D}_\text{val};\sigma)$}
    \end{subfigure}
    \begin{subfigure}[t]{0.32\textwidth}
        \centering
        \includegraphics[scale=0.37]{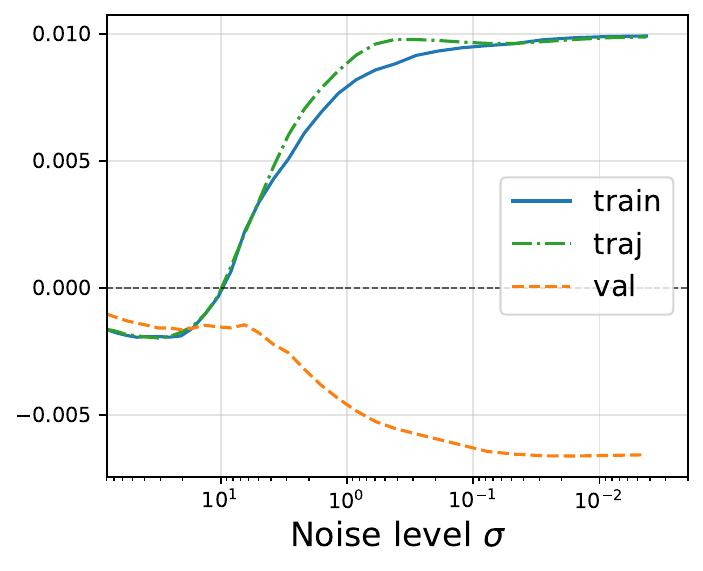}
        \caption{$B_{20}(Q; \sigma)$}
    \end{subfigure}
    \caption{The gap between the absolute similarity to the references between predictions from intermediate states of trajectories (green line) and predictions from noised data points (blue and orange lines) closes with increasing model size. All plots show fully trained models. \textbf{First column:} EDM2-XS. \textbf{Second column:} EDM2-XL.}
    \label{fig:in64-Q-v-sigma}
\end{figure}
\vspace*{\fill}
\vspace*{\fill}
\begin{figure}[h]
    \centering
    \begin{subfigure}[t]{0.32\textwidth}
        \centering
        \includegraphics[scale=0.37]{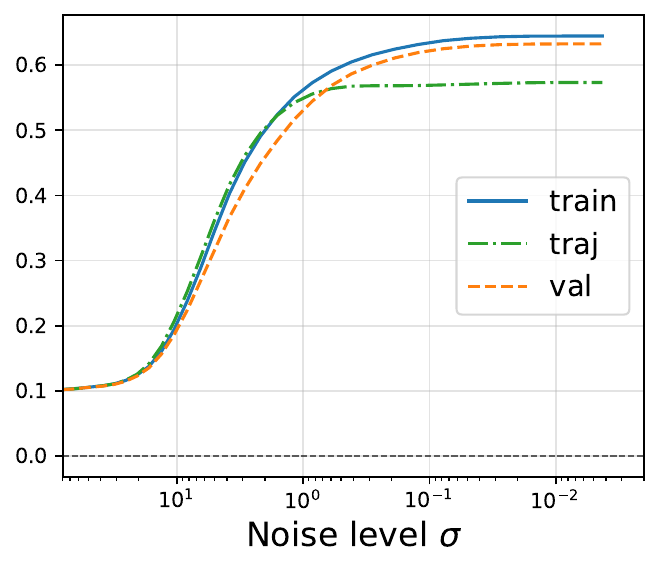}
        \caption{$S_{5}(Q, {\cal D}_\text{train};\sigma)$}
    \end{subfigure}
    \begin{subfigure}[t]{0.32\textwidth}
        \centering
        \includegraphics[scale=0.37]{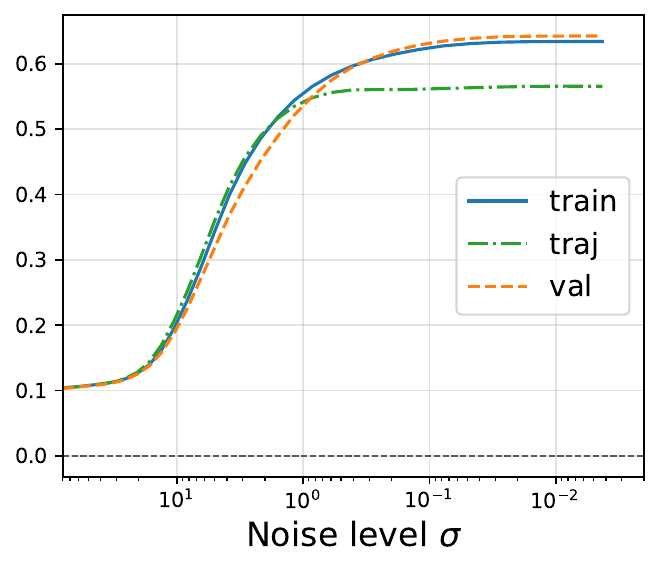}
        \caption{$S_{5}(Q, {\cal D}_\text{val};\sigma)$}
    \end{subfigure}
    \begin{subfigure}[t]{0.32\textwidth}
        \centering
        \includegraphics[scale=0.37]{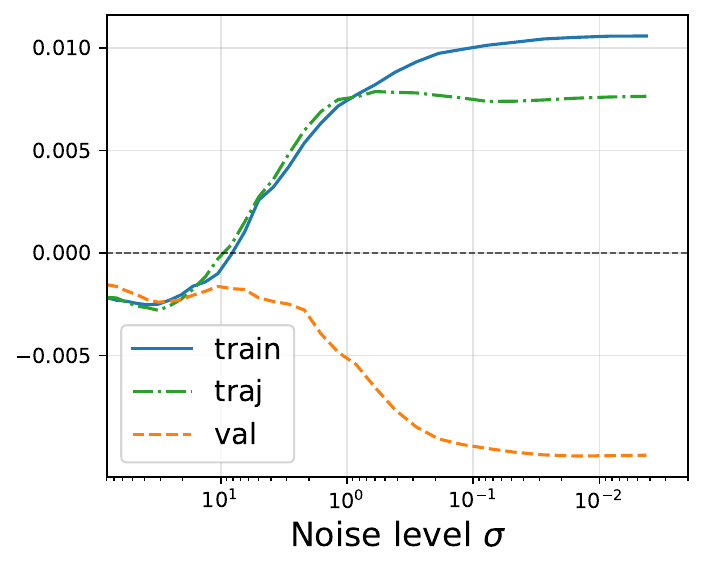}
        \caption{$B_{5}(Q; \sigma)$}
    \end{subfigure}
    
    % \vspace{3mm}
     
    % \begin{subfigure}[t]{0.32\textwidth}
    %     \centering
    %     \includegraphics[scale=0.37]{imgs/Q_vs_sigma_k10_in64_edm2-s_Sk_train_1073741.pdf}
    %     \caption{$k=10$}
    % \end{subfigure}
    % \begin{subfigure}[t]{0.32\textwidth}
    %     \centering
    %     \includegraphics[scale=0.37]{imgs/Q_vs_sigma_k10_in64_edm2-s_Sk_val_1073741.pdf}
    %     \caption{$k=10$}
    % \end{subfigure}
    % \begin{subfigure}[t]{0.32\textwidth}
    %     \centering
    %     \includegraphics[scale=0.37]{imgs/Q_vs_sigma_k10_in64_edm2-s_Bk_1073741.pdf}
    %     \caption{$k=10$}
    % \end{subfigure}
    
    % \vspace{3mm}
     
    % \begin{subfigure}[t]{0.32\textwidth}
    %     \centering
    %     \includegraphics[scale=0.37]{imgs/Q_vs_sigma_k25_in64_edm2-s_Sk_train_1073741.pdf}
    %     \caption{$k=25$}
    % \end{subfigure}
    % \begin{subfigure}[t]{0.32\textwidth}
    %     \centering
    %     \includegraphics[scale=0.37]{imgs/Q_vs_sigma_k25_in64_edm2-s_Sk_val_1073741.pdf}
    %     \caption{$k=25$}
    % \end{subfigure}
    % \begin{subfigure}[t]{0.32\textwidth}
    %     \centering
    %     \includegraphics[scale=0.37]{imgs/Q_vs_sigma_k25_in64_edm2-s_Bk_1073741.pdf}
    %     \caption{$k=25$}
    % \end{subfigure}

    \vspace{3mm}

    \begin{subfigure}[t]{0.32\textwidth}
        \centering
        \includegraphics[scale=0.37]{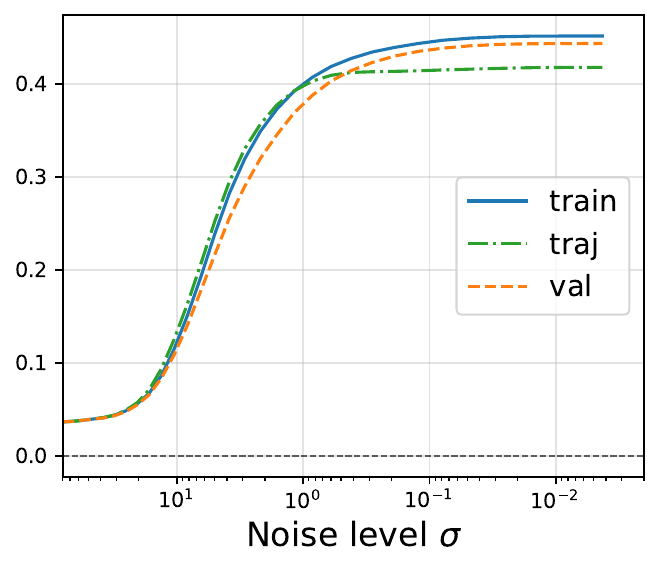}
        \caption{$S_{45}(Q, {\cal D}_\text{train};\sigma)$}
    \end{subfigure}
    \begin{subfigure}[t]{0.32\textwidth}
        \centering
        \includegraphics[scale=0.37]{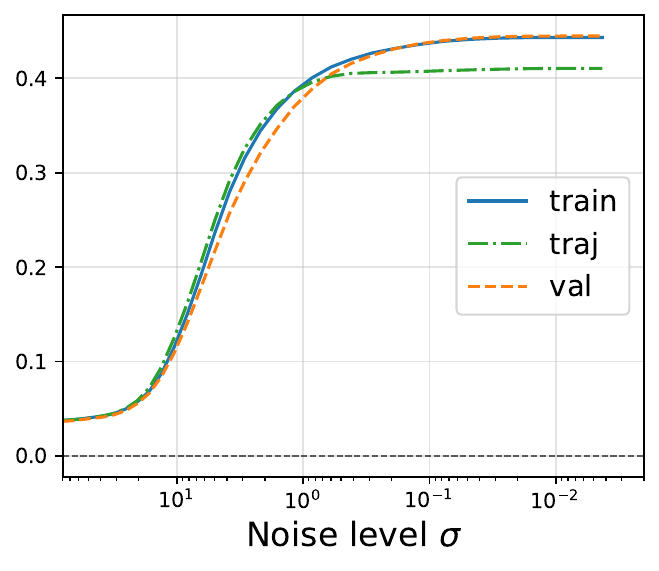}
        \caption{$S_{45}(Q, {\cal D}_\text{val};\sigma)$}
    \end{subfigure}
    \begin{subfigure}[t]{0.32\textwidth}
        \centering
        \includegraphics[scale=0.37]{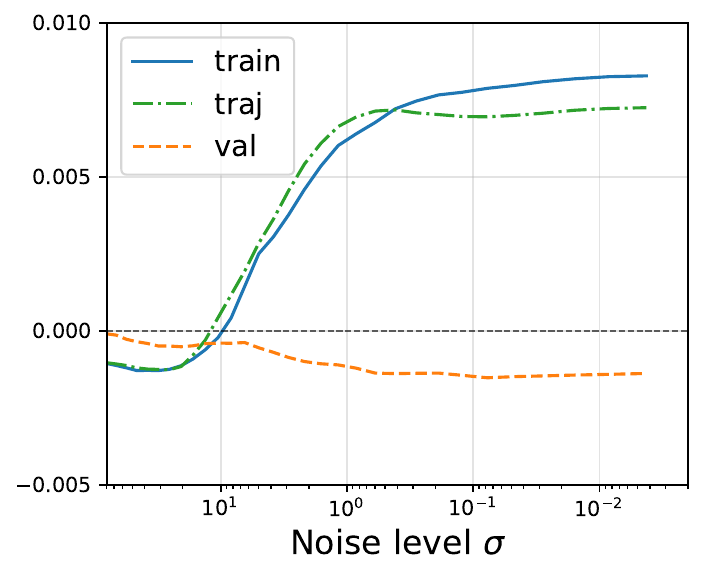}
        \caption{$B_{45}(Q; \sigma)$}
    \end{subfigure}
    \caption{Qualitative trends in the metrics of \cref{sec:inference} are robust to the choice of $k$. All plots show fully trained models.}
    \label{fig:in64-Q-v-sigma-k}
\end{figure}
\vspace*{\fill}
\begin{figure}[h]
    \centering
    \begin{subfigure}[t]{0.45\textwidth}
        \centering
        \includegraphics[scale=0.5]{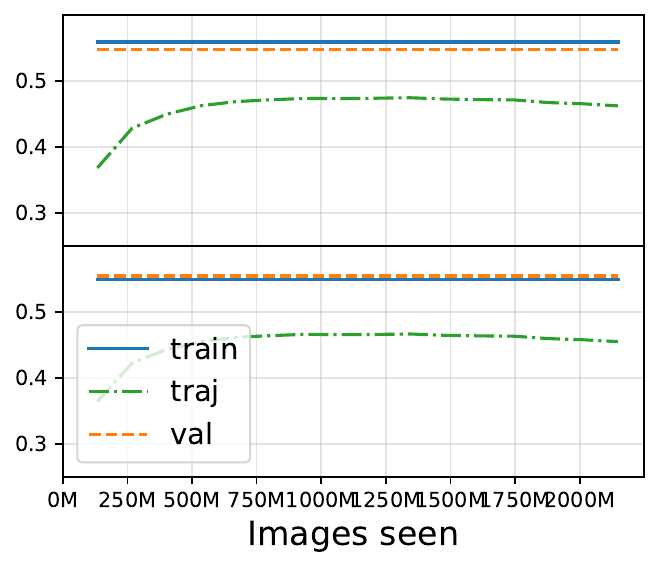}
        \caption{$S_{20}(Q, {\cal D}_\text{train};\sigma)$}
    \end{subfigure}
    \begin{subfigure}[t]{0.45\textwidth}
        \centering
        \includegraphics[scale=0.5]{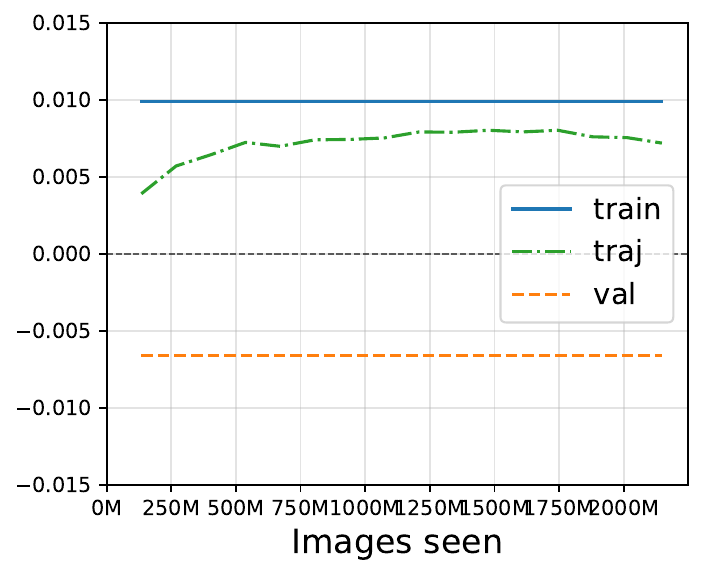}
        \caption{$S_{20}(Q, {\cal D}_\text{train};\sigma)$}
    \end{subfigure}

    \vspace{3mm}

   \begin{subfigure}[t]{0.45\textwidth}
        \centering
        \includegraphics[scale=0.5]{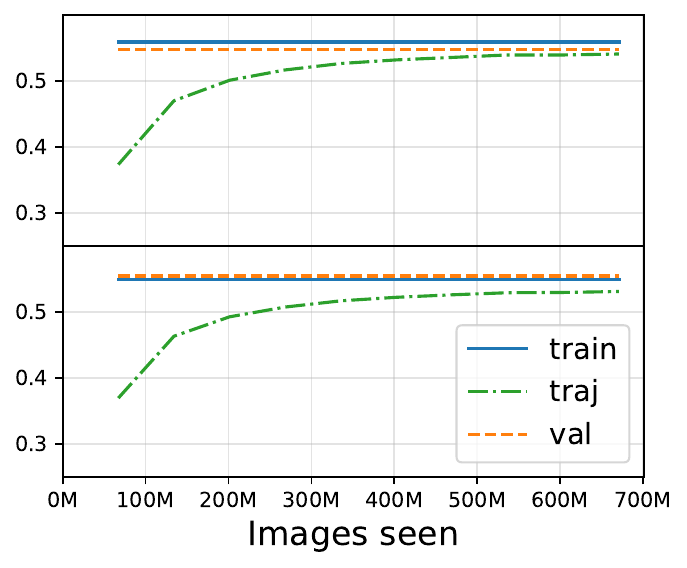}
        \caption{$S_{20}(Q, {\cal D}_\text{train};\sigma)$}
    \end{subfigure}
    \begin{subfigure}[t]{0.45\textwidth}
        \centering
        \includegraphics[scale=0.5]{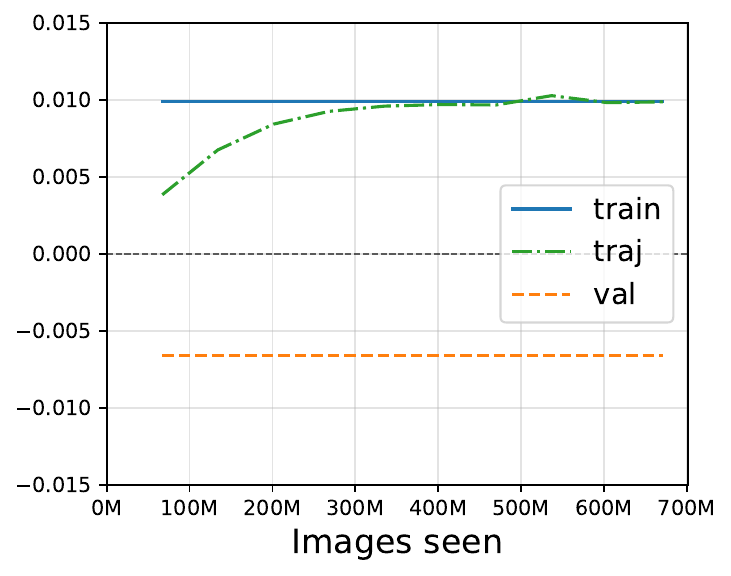}
        \caption{$S_{20}(Q, {\cal D}_\text{train};\sigma)$}
    \end{subfigure}
    
    \caption{The gap between the absolute similarity to the references between predictions from intermediate states of trajectories (green line) and predictions from noised data points (blue and orange lines) closes with increasing model size. All plots show fully trained models. \textbf{First column:} EDM2-XS. \textbf{Second column:} EDM2-XL.}
    \label{fig:in64-Q-v-ckpt}
\end{figure}
\vspace*{\fill}

\end{document}